\documentclass[journal]{IEEEtran}

\ifCLASSINFOpdf
   \usepackage[utf8]{inputenc}
   \usepackage[pdftex]{graphicx}
   \usepackage{amsmath,epsfig,amssymb,subfigure,bm,dsfont}
   \usepackage{multirow}
   \usepackage{epstopdf}
   \usepackage{tabu}
   \usepackage{cite}
   \usepackage{color}
   \usepackage{balance}
   \usepackage[citecolor=blue, colorlinks]{hyperref}
\else

\fi

\begin{document}

\title{Deep Posterior Distribution-based Embedding for Hyperspectral Image Super-resolution}

\author{Jinhui Hou, Zhiyu Zhu, Junhui Hou,~\emph{Senior Member, IEEE},~
Huanqiang Zeng,~\emph{Senior Member, IEEE}, \\Jinjian Wu, \textit{Member, IEEE}, and Jiantao Zhou,~\emph{Senior Member, IEEE}
\thanks{This work was supported in part by the Hong Kong Research Grants Council under
Grants 11219019 and 11218121, and in part by the Basic
Research General Program of Shenzhen Municipality under Grant JCYJ20190808183003968. \textit{Jinhui Hou and Zhiyu Zhu contributed to this work equally. Corresponding author: Junhui Hou}}

\thanks{J. Hou, Z. Zhu, and J. Hou are with the Department of Computer Science, City University of Hong Kong, Hong Kong, and also with the City
University of Hong Kong Shenzhen Research Institute, Shenzhen 518057,
China (e-mail: jhhou3-c@my.cityu.edu.hk; zhiyuzhu2@my.cityu.edu.hk; jh.hou@cityu.edu.hk).}%
\thanks{H. Zeng is with the School of Engineering and School of
Information Science and Engineering, Huaqiao University, Xiamen, China (e-mail: zeng0043@hqu.edu.cn).}
\thanks{J. Wu is with the School of Artificial Intelligence, Xidian University
Xi’an, China (e-mail: jinjian.wu@mail.xidian.edu.cn).}
\thanks{J. Zhou is with the Department of Computer and Information Science,
University of Macau, Macau (e-mail: jtzhou@um.edu.mo).}
}

\markboth{
}
{Shell \MakeLowercase{\textit{et al.}}: Bare Demo of IEEEtran.cls for IEEE Journals}

\maketitle

\begin{abstract}
In this paper, we investigate the problem of hyperspectral (HS) image spatial super-resolution via deep learning.
Particularly, we focus on how to embed the high-dimensional spatial-spectral information of HS images efficiently and effectively.
Specifically, in contrast to existing methods adopting empirically-designed network modules,
we formulate HS embedding as an approximation of the posterior distribution of a set of carefully-defined HS embedding events, including layer-wise spatial-spectral feature extraction and network-level feature aggregation. Then, we incorporate the proposed feature embedding scheme into a source-consistent super-resolution  framework that is physically-interpretable, producing PDE-Net, in which high-resolution (HR) HS images are iteratively refined from the residuals between input low-resolution (LR) HS images and pseudo-LR-HS images degenerated from reconstructed HR-HS images via  probability-inspired HS embedding.
Extensive experiments over three common benchmark datasets demonstrate that PDE-Net achieves superior performance over state-of-the-art methods.
Besides, the probabilistic characteristic of this kind of networks can provide the epistemic uncertainty of the network outputs,
which may bring additional benefits when used for other HS image-based applications.
The code will be publicly available at \url{https://github.com/jinnh/PDE-Net}.

\end{abstract}

\begin{IEEEkeywords}
Hyperspectral imagery, deep learning, super-resolution, convolution, high-dimensional feature extraction, probability.
\end{IEEEkeywords}

\IEEEpeerreviewmaketitle

\section{Introduction}

\IEEEPARstart
{H}{yperspectral} (HS) imaging aims to capture the continuous electromagnetic spectrum of real-world scenes/objects. Benefiting from such dense spectral resolution, HS images are widely applied in numerous areas, such as agriculture \cite{park2015hyperspectral,lu2020recent}, military \cite{shimoni2019hypersectral,jia2020status}, and environmental monitoring \cite{banerjee2020uav,mishra2017close}.
Unfortunately, due to limited sensor resolution, it's hard to acquire HS images with both high spatial and spectral resolution via single-shot HS imaging devices. The inevitable trade-off between the spatial and spectral resolution results in much lower spatial resolution than traditional RGB images, which may limit the performance of downstream HS image-based applications.

\begin{figure}[!t]
\centering
\includegraphics[width=0.95\linewidth]{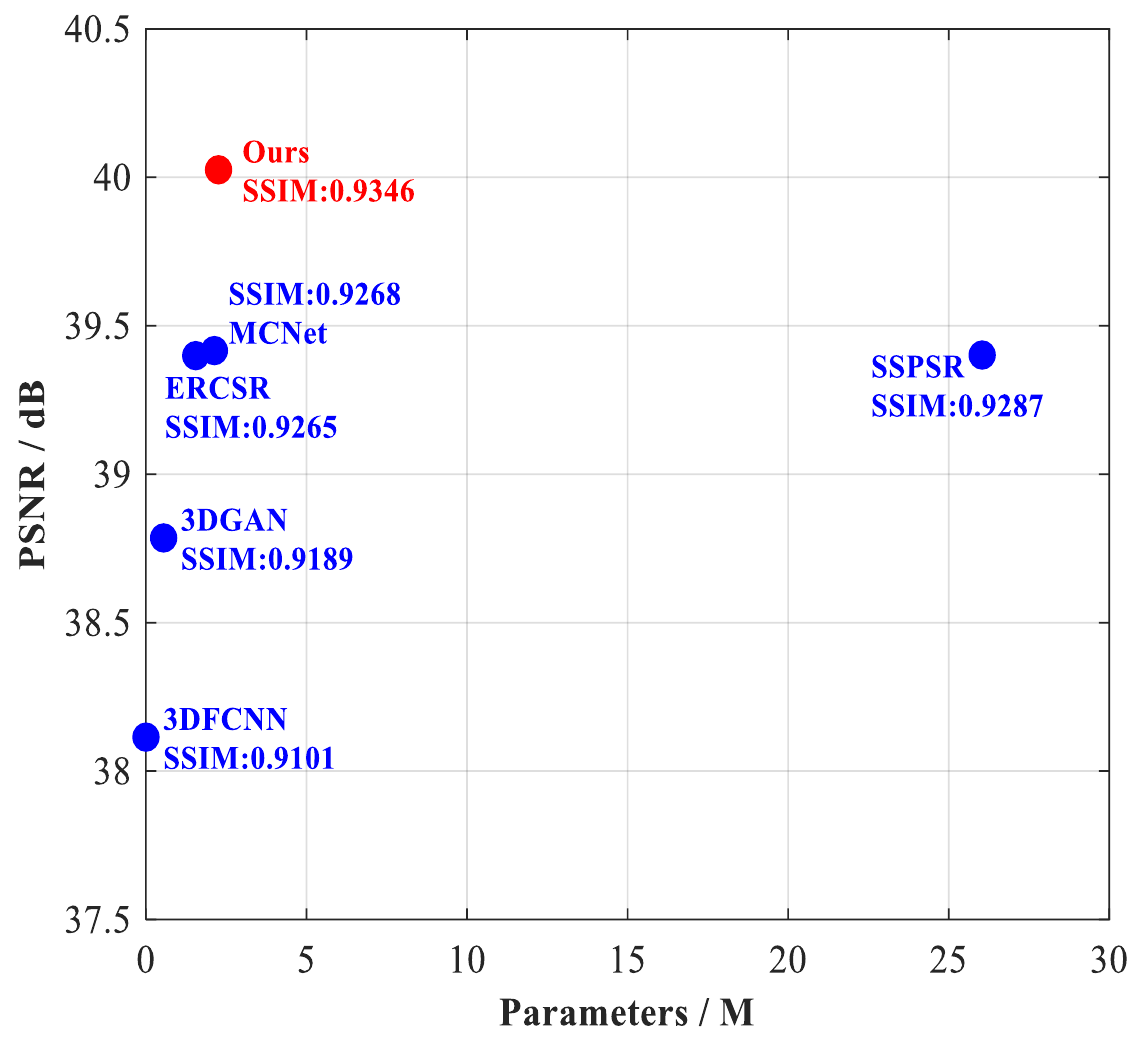}
\caption{Comparison of the number of parameters and average reconstruction quality (PSNR/SSIM) of our PDE-Net and  state-of-the-art deep learning-based methods, including 3DFCNN \cite{Mei2017Hyperspectral}, 3DGAN \cite{Li2020Hyperspectral}, ERCSR \cite{Li2021Exploring} , MCNet \cite{Li2020Mixed}, and SSPSR \cite{Jiang2020Learning}, on the Harvard dataset for $4\times$ super-resolution.
}
\label{fig:params-psnr}
\end{figure}

Instead of relying on the development of hardware,  computational methods known as super-resolution have been proposed for reconstructing high-resolution (HR) HS images from low-resolution (LR) ones
\cite{Jiang2020Learning,Dong2016Hyperspectral,yi2018hyperspectral,Mei2017Hyperspectral,Li2020Hyperspectral,Li2020Mixed,Li2021Exploring}.
Specifically, the early works explicitly formulate HS image super-resolution as constrained optimization problems regularized by prior knowledge,
such as, sparsity \cite{Akhtar2015Bayesian,xu2019nonlocal,Han2018Self}, non-local similarity \cite{Dian2017Hyperspectral,Dian2019Learning}, and low-rankness \cite{dian2019hyperspectral,xue2021spatial}. Besides, auxiliary information \cite{Dian2019Learning,Dong2016Hyperspectral}, \cite{vicinanza2014pansharpening,fei2019convolutional}, e.g., HR RGB and panchromatic images, was incorporated to improve reconstruction quality. However,
the limited representation ability of these optimization-based methods is insufficient to model such a severely ill-posed problem,
making the quality of reconstructed HR-HS images still unsatisfied.
Owing to the powerful representational ability, recent deep learning-based HS image super-resolution methods have improved the reconstruction quality significantly
\cite{Xie2019Multispectral,yao2020cross,qu2018unsupervised,Zhu2021Hyperspectral,Mei2017Hyperspectral,Li2020Mixed,Jiang2020Learning,zhu2021deep,zhu2021semantic}. For deep learning-based HS image super-resolution, one of the critical issues is how to effectively and efficiently extract/embed the high-dimensional spatial-spectral information.
Most of the existing methods design the feature extraction/embedding module by empirically combining some common convolutions in the dense or residual fashion, such as separately convolving on spatial and spectral domains \cite{Li2021Exploring,Xie2018Rethinking}, directly utilizing 3D convolution \cite{Mei2017Hyperspectral}, or using both 2D and 3D convolutional layers \cite{Li2020Mixed}, such network architectures may be not optimal, thus compromising performance.

In contrast to existing methods that adopt empirically-designed convolutional modules to embed the high-dimensional spatial-spectral information of HS images,
we propose to cast this process as an approximation to the posterior distribution of a set of carefully-defined HS embedding events, including layer-wise spatial-spectral feature extraction and network-level feature aggregation. Then, we incorporate the proposed feature embedding scheme into a source-consistent spatial super-resolution framework that is built upon the degradation process of LR-HS images from HR-HS ones and thus physically-interpretable, leading to PDE-Net, where a coarse HR-HS image is first initialized and then iteratively  refined  by learning residual maps from the differences between the input LR-HS image and the pseudo-LR-HS image re-degenerated from the reconstructed HR-HS image.
Extensive experiments on three common benchmark datasets demonstrate the significant superiority of the proposed PDE-Net over multiple state-of-the-art methods.

In summary, our contributions are two-fold:
\begin{itemize}

\item we formulate the embedding of the high-dimensional spatial-spectral information of HS images from the probabilistic perspective and propose a generic HS feature embedding scheme; and

\item we incorporate the proposed feature embedding scheme into a physically-interpretable deep framework to construct an end-to-end HS image super-resolution method and experimentally demonstrate its advantages over state-of-the-art ones. Besides, the probabilistic characteristic of the method can bring additional benefits, e.g., the uncertainty of outputs.

\end{itemize}

The rest of this paper is organized as follows. Section~\ref{sec:Re} briefly reviews existing works. Section~\ref{sec:proposed} presents the proposed framework in detail, followed by extensive experiments and analysis in  Section~\ref{sec:experiments}.
Finally, Section~\ref{sec:con} concludes this paper.

\section{Related Work}
\label{sec:Re}

\subsection{Single HS Image Super-resolution}
The early works explicitly  formulate  HS  image  super-resolution  as  constrained optimization  problems, in which some priors are explored to regularize the solution space.  For example, Wang \emph{et al.} \cite{Wang2017Hyperspectral}  modeled the three characteristics of HS images, i.e., the global correlation in the spectral domain, the non-local self-similarity in the spatial domain, and the local smooth structure across both spatial and spectral domains. Huang \emph{et al.} \cite{huang2014super} utilized the low-rank and group-sparse modeling to spatially super-resolve single HS images. Zhang \emph{et al.} \cite{zhang2012super} proposed a maximum a posterior-based HS image super-resolution algorithm.
Recently, many deep learning-based methods for single HS image super-resolution have been proposed, which improve the reconstruction quality of traditional optimization-based methods dramatically. For example,
Yuan \emph{et al.} \cite{Yuan2017Hyperspectral} designed a transfer learning model to recover HR-HS image by utilizing the knowledge from the natural image and enforcing collaborations between LR and HR-HS images via non-negative matrix factorization. Li \emph{et al.} \cite{Li2018Single} proposed a grouped deep recursive residual network
with a grouped recursive module embedded to effectively formulate the ill-posed 
mapping function from LR- to HR- HS images.
To simultaneously explore spatial and spectral information, Hu \emph{et al.} \cite{Hu2020Hyperspectral} proposed an intrafusion network to jointly learn the spatial information, spectral information, and spectral difference. Jiang \emph{et al.} \cite{Jiang2020Learning} designed a spatial-spectral prior network with progressive upsampling and grouped convolutions with shared parameters.
Mei \emph{et al.} \cite{Mei2017Hyperspectral} proposed a 3D full convolution neural network (CNN) to well explore both the spatial context and spectral correlation.
Li \textit{et al}. \cite{Li2020Hyperspectral} presented a 3D generative adversarial network with a band attention mechanism to alleviate the spectral distortion problem.
However, 3D convolution is usually with high computational and memory complexity. Inspired by the separable 3D CNN model \cite{Xie2018Rethinking}, Li \emph{et al.} \cite{Li2020Mixed} proposed a mixed convolutional module, including 2D convolution and separable 3D convolution, to fully extract spatial and spectral features. In \cite{Li2021Exploring},
the relationship between 2D and 3D convolution was explored to achieve HS image super-resolution.

Although various network architectures/convolutions were designed to fully and efficiently exploit the high-dimensional spatial-spectral information for achieving high reconstruction quality, they were empirically designed based on human knowledge, which may be not optimal, thus limiting performance.

\begin{figure*}[!t]
\centering
\includegraphics[width=0.9\linewidth]{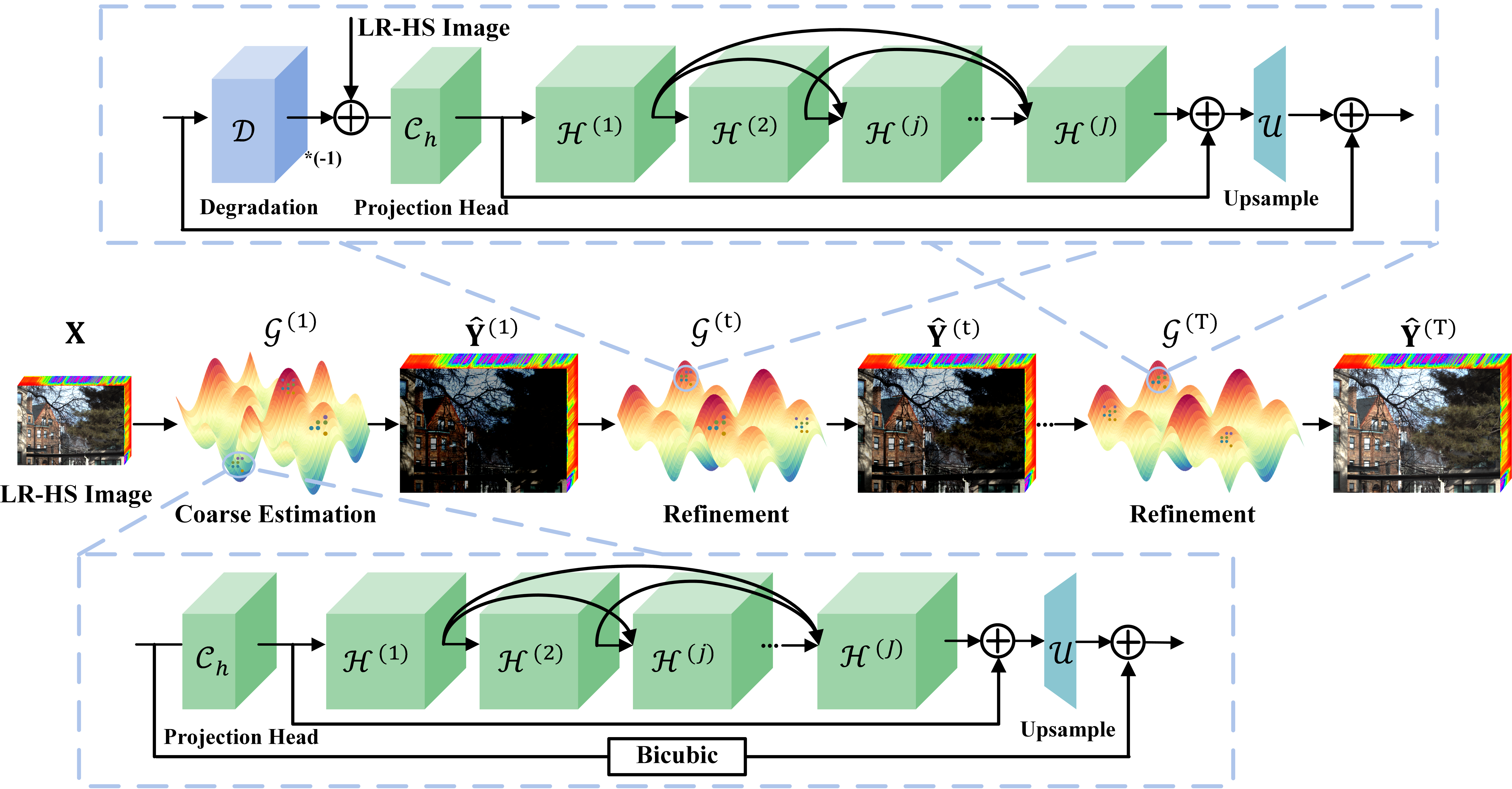}
\caption{Illustration of the flowchart of the proposed PDE-Net for HS image super-resolution. Our PDE-Net consists of  coarse estimation and multi-stage source-consistent HS refinement. More importantly, PDE-Net adaptively learns the architecture of the basic HS embedding unit ($\mathcal{H}^{(j)}$) as well as the connection between different units from the probabilistic perspective, which is fundamentally different from existing methods with empirically-designed architectures.
}
\label{fig:hsisr-framework}
\end{figure*}

\subsection{Fusion-based HS Image Super-resolution}

Different from single HS image super-resolution, fusion-based HS image super-resolution methods employ additional data, e.g.,  HR RGB images, to improve performance.
Many traditional methods have been proposed, such as Bayesian inference-based \cite{Akhtar2015Bayesian, Akhtar2016Hierarchical}, matrix factorization-based \cite{Lanaras2015Hyperspectral, Dian2017Hyperspectral, Dian2019Learning}, and sparse representation-based \cite{Akhtar2014Sparse, Dong2016Hyperspectral, Han2018Self}. To be specific, under the assumption that each spectrum can be linearly represented with  multiple spectral atoms, Dian \emph{et al.} \cite{Dian2019Learning} proposed a matrix factorization-based approach. Han \emph{et al.} \cite{Han2018Self} designed a self-similarity constrained sparse representation approach to form the global-structure groups and local-spectral super-pixels.
The recent deep learning-based methods \cite{Xie2019Multispectral, Wang2019Deep, Zhang2020Unsupervised, Zhu2021Hyperspectral} improve the performance of fusion-based HS image super-resolution significantly. For example, Xie \emph{et al.} \cite{Xie2019Multispectral} constructed a deep network, which mimics the iterative algorithm for solving the explicitly formed fusion model, to merge an HR multispectral image and an LR-HS image to generate an HR-HS image. Wang \emph{et al.} \cite{Wang2019Deep} proposed a deep blind iterative fusion network to iteratively optimize the estimation of the observation model and fusion process. Zhu \emph{et al.} \cite{Zhu2021Hyperspectral} designed a progressive zero-centric residual network with the spectral-spatial separable convolution to enhance the performance of HS image reconstruction.

Despite fusion-based methods have achieved remarkable performance, the above-mentioned methods highly rely on the additional co-registered HR images, which may be difficult to obtain. Recently, to tackle the registration challenge,
Qu \emph{et al.} \cite{Qu2022Unsupervised} presented an registration-free and unsupervised mutual Dirichlet-Net, namely $u^2$-MDN.

\section{Proposed Method}
\label{sec:proposed}

\subsection{Problem Statement and Overview}
\label{sec:Pfam}
Given an LR-HS image denoted as $\mathbf{X}$ $\in\mathbb{R}^{B\times hw}$ with $h \times w$ being the spatial dimensions and $B$ being the number of spectral bands, we aim to recover an  HR-HS image denoted as $\mathbf{Y}$ $\in\mathbb{R}^{B\times HW}$ ($H=\alpha h$ and $W=\alpha w$ where $\alpha>1$ is the scale factor).
The degradation process of $\mathbf{X}$ from $\mathbf{Y}$ can be generally written as
\begin{equation}
\label{equ:1}
\mathbf{X} = \mathbf{YD} + \mathbf{N}_{z},
\end{equation}
where $\mathbf{D}$ $\in\mathbb{R}^{HW\times hw}$ is the degeneration matrix composed of the blurring and down-sampling operators
and $\mathbf{N}_{z}\in \mathbb{R}^{B\times hw}$ stands for the noise. To tackle such an ill-posed reconstruction problem, inspired by the great success of deep CNNs in image/video processing applications, we will consider a deep learning-based framework named PDE-Net, as illustrated in Fig. \ref{fig:hsisr-framework}.
Note that instead of designing a new overall framework to achieve performance improvement, we focus on the efficient and effective feature embedding manner for capturing the high-dimensional characteristics of HS images. To be specific, motivated by iterative back-projection refinement works \cite{romano2015boosting,Tao2017Zero,Wang2019Deep},
we propose a source-consistent reconstruction framework, in which a coarse HR-HS image is first initialized and then iteratively refined by learning residual maps from the differences between the input LR-HS image and the pseudo-LR-HS image re-degenerated from the reconstructed HR-HS image. More importantly, to explore the high-dimensional spatial-spectral information of HS images efficiently and effectively, we propose posterior distribution-based HS embedding, the core module of our PDE-Net for feature embedding, which models the process of embedding HS images as an approximation of posterior distributions.
Owing to the explicit problem formulation,  the proposed PDE-Net is physically-interpretable and compact.

In what follows, we will first provide the overall framework before presenting the proposed probability-based feature embedding scheme.

\subsection{Source-consistent Reconstruction}

According to Eq. (\ref{equ:1}), it can be deduced that if the reconstructed HR-HS image $\widehat{\mathbf{Y}}\in\mathbb{R}^{B\times HW}$ via a typical method approximates $\mathbf{Y}$ well, the re-degenerated LR-HS image $\widehat{\mathbf{X}}\in\mathbb{R}^{B\times hw}$ from $\widehat{\mathbf{Y}}$ via Eq. (\ref{equ:1}) should be very close to $\mathbf{X}$.
Equivalently, the difference between $\widehat{\mathbf{X}}$ and $\mathbf{X}$ indicates the deviation of $\widehat{\mathbf{Y}}$ from $\mathbf{Y}$.
Based on this deduction, as illustrated in Fig. \ref{fig:hsisr-framework}, we propose a  source-consistent reconstruction framework, composed of two modules, i.e., coarse estimation and iterative refinement.

\subsubsection{Coarse estimation} In this module, we estimate a coarse HR-HS image denoted as  $\widehat{\mathbf{Y}}^{(1)}\in\mathbb{R}^{B\times HW}$ from $\mathbf{X}$ in a residual learning manner, i.e.,
\begin{equation}
{\widehat{\mathbf{Y}}^{(1)}} = \mathcal{G}^{(1)}\left(\mathbf{X}\right) + \mathcal{I}\left(\mathbf{X}\right),
    \label{con:eqution3}
\end{equation}
where $\mathcal{G}^{(1)}(\cdot)$ stands for the process of regressing residuals from its input,
the details of which are provided in Section \ref{Sec:PDHSE}, and $\mathcal{I}(\cdot)$ denotes the bicubic interpolation operator.

\subsubsection{Iterative refinement} Let $\mathcal{D}(\cdot)$ be a single convolutional layer with the stride equal to
$\alpha$ to mimic the degradation process in Eq. (\ref{equ:1}), i.e., $\widehat{\mathbf{X}}=\mathcal{D}(\widehat{\mathbf{Y}})$, and the kernel size is set to $5$ (resp. 9) when $\alpha$ = 4 (resp. 8). We design a multi-stage structure to iteratively refine the coarse estimation by exploring the differences between $\mathbf{X}$ and $\widehat{\mathbf{X}}$, and at the $t$-th ($t=2,\cdots,T$) stage, the refinement process is written as
\begin{equation}
{\widehat{\mathbf{Y}}^{(t)}} = \mathcal{G}^{(t)}\left(\mathbf{X}- \mathcal{D}\left(\widehat{\mathbf{Y}}^{(t-1)}\right)\right) + \widehat{\mathbf{Y}}^{(t-1)},
    \label{con:eqution4}
\end{equation}
where $\mathcal{G}^{(t)}(\cdot)$ is the set of HS embedding events involved in the $t$-th stage.

\subsection{Posterior Distribution-based HS Embedding}
\label{Sec:PDHSE}

Learning representative embeddings from high-dimensional HS images is a crucial issue for deep learning-based HS image processing methods.
As an HS image is a 3D cube, the 3-D convolution is an intuitive choice for feature extraction, which has demonstrated its effectiveness \cite{Mei2017Hyperspectral}. However, compared with 1-D and 2-D convolutions, the 3-D convolution results in a significant increase of network parameters, which may potentially cause over-fitting and consumption of huge computational resources. By analogy with the approximation of a high-dimensional filter with multiple low-dimensional filters in the field of signal processing,
one can perform multiple low-dimensional convolutions along one or two out of three dimensions separately, and then aggregate them together to cover all the three dimensions.
However, some questions are naturally posed: `` (1) how to select convolutional patterns? and (2) how to effectively and efficiently aggregate those convolutional layers together?"   Based on human prior knowledge, previous works \cite{dong2019deep}, \cite{wang2020spatial}, \cite{Li2020Mixed}, \cite{Li2021Exploring} empirically combine some low-dimensional convolutional layers,  such as 1-D convolutions in the spectral dimension and  2-D convolution in the spatial dimension, which maybe not optimal, thus compromising performance.

By contrast, from the probabilistic view,
we formulate HS embedding as to optimize the distribution $\mathcal{P}(\widehat{\mathbf{Y}}|\mathbf{X},\mathbb{T})$, where
$\mathbb{T}=\{(\mathbf{X}_v, \mathbf{Y}_v)\}_{v=1}^V$ is a set of paired training samples. Let $\mathcal{G}=\{\mathcal{G}^{(t)}\}_{t=1}^T$ be the set of feasible events for HS embedding at each stage, including network architectures and corresponding weights. With the Bayesian theorem,
we can rewrite this process as
\begin{equation}
\label{equ:gdprogress}
\mathcal{P}(\widehat{\mathbf{Y}}|\mathbf{X},\mathbb{T}) = \int \mathcal{P}(\widehat{\mathbf{Y}}|\mathbf{X},\mathcal{G}) \mathcal{P}(\mathcal{G}|\mathbb{T}) ~~ d\mathcal{G},
\end{equation}
where $\mathcal{P}(\widehat{\mathbf{Y}}|\mathbf{X},\mathcal{G})$ is the model likelihood, which could be calculated via a single inference process,
and the posterior distribution $\mathcal{P}(\mathcal{G}|\mathbb{T})$ captures the distribution of a set of plausible models for the dataset $\mathbb{T}$.
Thus, to achieve HS embedding, we can optimize a distribution $\mathcal{Q}(\mathcal{G})$ to approximate the intractable posterior distribution $\mathcal{P(\mathcal{G}|\mathbb{T})}$.

\label{ssec:ssfs}

Specifically, to model the distribution $\mathcal{Q}(\mathcal{G})$,
we first define the set of plausible HS embedding events $\mathcal{G}$.
Let $\mathcal{G}^{(t)} = \left\{ \{(\mathbf{K}^{(j)},\mathcal{H}^{(j)})\}_{j=1}^J,
~\mathcal{C}_h,~\mathcal{U}\right\}$, where $\mathbf{K}^{(j)} \in \{0,1\}^{j-1}$ is a binary vector of length $j-1$,
indicating whether the features from previous $j-1$ units are used (i.e., if the typical element of $\mathbf{K}^{(j)}$ is equal to 1, the features of the corresponding unit will be used), $\mathcal{H}^{(j)}$ stands for a unit to extract high-level HS features, $\mathcal{C}_h(\cdot)$ is a convolutional layer as an projection head to lift up the number of channel for input feature maps, and $\mathcal{U}(\cdot)$ is a spatial upsampling layer for transforming the LR-feature maps to an HR-HS image.
Moreover, to handle the high-dimensionality of HS images efficiently, we further introduce spatial and spectral separable convolutional layers
for local HS feature extraction,  i.e., $\mathcal{H}^{(j)} = \left\{ \mathbf{L}^{(j)},~\mathcal{C}_{spe}^{(j)}(\cdot),~  \mathcal{C}_{spa}^{(j)}(\cdot) \right\}$, where $\mathcal{C}_{spe}^{(j)}(\cdot)$ and $\mathcal{C}_{spa}^{(j)}(\cdot)$ denote convolutional layers in  spectral and spatial domains, respectively, and
$\mathbf{L}^{(j)} \in \{0,1\}^{2}$
is a binary vector of length 2, indicating whether the spatial or spectral embedding layer is used (i.e., if the typical element of $\mathbf{L}^{(j)}$ is equal to 1, the corresponding layer will be used). We name the network built upon $\mathcal{G}^{(t)}$ with  all elements of $\{\mathbf{K}^{(j)}\}_{j=1}^{J}$ and $\{\mathbf{L}^{(j)}\}_{j=1}^{J}$ fixed to 1 the template network (Template-Net). Next, we will demonstrate the approach to learn the distribution $\mathcal{Q}(\mathcal{G})$.

\begin{figure}
    \centering
    \includegraphics[width=0.75\linewidth,height=0.8\linewidth]{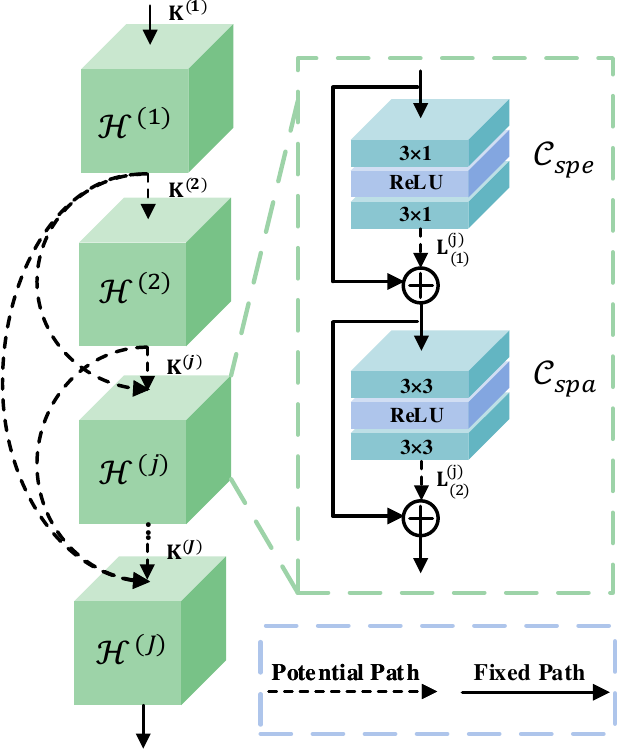}
    \caption{Illustration of a set of possible aggregation unit $\mathcal{H}^{(j)}$ for HS image embedding. \textbf{Left}: the network-level aggregation pattern controlled by $\mathbf{K}^{(j)}$; \textbf{Right:} the feasible feature embedding patterns in the unit $\mathcal{H}^{(j)}$ controlled by $\mathbf{L}^{(j)}$.}
    \label{fig:Aggregation}
\end{figure}

Considering that introducing the dropout operation into CNNs could objectively minimize the Kullback–Leibler divergence between an approximate distribution $\mathcal{Q}(\mathcal{G})$ and the model posterior $\mathcal{P}(\mathcal{G}|\mathbb{T})$ \cite{gal2016dropout}, we model the distribution $\mathcal{Q}(\mathcal{G})$ via training a template network whose binary vectors
are replaced with variables following independent learnable Bernoulli distributions. Specifically, as shown in Fig. \ref{fig:Aggregation}, both
the path for feature aggregation ($\{\mathbf{K}^{(j)}\}_{j=1}^{J}$) and the local feature embedding pattern ($\{\mathbf{L}^{(j)}\}_{j=1}^{J}$) are replaced with masks of logits $\epsilon \sim \mathcal{B}(p)$, where $\mathcal{B}(p)$ denotes the Bernoulli distribution with probability $p$. However, the classic sampling process is hard to manage a differentiable linkage between the sampling results and the probability, which restricts the gradient descent-based optimization process of CNNs. Besides, such dense aggregations in CNNs will result in a huge number of feature embeddings, which makes the CNNs hard to be optimized. Thus, we also need to explore an efficient and effective way for aggregating those features masked by binary logits. In what follows, we discuss how to deal with the  two aspects.

To obtain a differentiable sampling manner of logits $\epsilon$, we use the Gumbel-softmax \cite{jang2016categorical} to relax the discrete Bernoulli distribution to continuous space. Mathematically, we formulate this process as
\begin{align}
    \small
\mathcal{M}(p) =  \mathsf{Sigmoid} \{ \frac{1}{\tau}& \left(\log p - \log (1 - p) +
\notag \right. \\
  &\left.
  \log (\log ({r_1})) - \log (\log ({r_2})) \right) \},
    \label{con:eqution3}
\end{align}
where $\mathsf{Sigmoid}(\cdot)$ refers to the sigmoid function; $r_1$ and $r_2$ are random noise with the standard uniform distribution in the range of $[0,1]$; $p$ is a learnable parameter encoding the probability of aggregations in the neural network; and the temperature $\tau > 0$ controls the similarity between $\mathcal{M}(p)$ and $\mathcal{B}(1-p)$, i.e.,
as $\tau \to 0$, the distribution of $\mathcal{M}(p)$ approaches $\mathcal{B}(1-p)$; while as $\tau \to \infty$, $\mathcal{M}(p)$ tends to be  a uniform distribution.

To aggregate the features efficiently and effectively, we design the network architecture at both network and layer levels.
Specifically, according to  Eq. (\ref{con:eqution3}), we approximate the discrete variable $\mathbf{K}^{(j)}$ by applying Gumbel-softmax $\mathcal{M}(\cdot)$ to continuous learnable variables $\widetilde{\mathbf{K}}^{(j)}$. Thus, we could formulate network-level feature aggregation as
\begin{align}
&\widetilde {\mathbf{F}}^{(j)} = { \mathcal{C}_{1 \times 1}}\left( {  { \mathbf{T}^{(0,j)} , \cdots, \mathbf{T}^{(j-1,j)}}} \right),\nonumber\\
&{\rm with} ~~ \mathbf{T}^{(k,j)} = {\mathbf{F}}^{(k)} \times \mathcal{M}\left(\widetilde{\mathbf{K}}^{(j)}_{(k)}\right),
    \label{con:eqution4}
\end{align}
where $\widetilde{\mathbf{F}}^{(j)} ~~ (1 \leq j \leq J)$ denotes the aggregated feature which would be fed into $\mathcal{H}^{(j)}$; $\mathbf{F}^{(k)} ~~ (1 \leq k \leq j-1)$ is the feature from the ${k}$-th embedding unit $\mathcal{H}^{(k)}$; $\mathbf{F}^{(0)}$ denotes an HS embedding extracted from the input of $\mathcal{H}^{(t)}(\cdot)$ by a single linear convolutional layer;
$\widetilde{\mathbf{K}}^{(j)}_{(k)}$ indicates the ${k}$-th element of the vector $\widetilde{\mathbf{K}}^{(j)}$ in a range of $[0,1]$, according to its meaning of the sampling probability of $\mathbf{K}^{(j)}$;
and $\mathcal{C}_{1 \times 1}(\cdot)$ is a $1 \times 1$ kernel to compress the feature embedding and activate them with the rectified linear unit (ReLU).

In analogy to the network-level design, we also introduce the continuous learnable weights $\widetilde{\mathbf{L}}^{(j)}$ with Gumbel-softmax to approximate the Bernoulli distribution of $\mathbf{L}^{(j)}$ in each feature embedding unit as
\begin{align}
&{\mathbf{F}^{(j + 1)}} = \mathbf{O}^{(j)} + {\mathcal{C}_{spa}^{(j)}}\left( \mathbf{O}^{(j)} \right) \times \mathcal{M}\left(\widetilde{\mathbf{L}}^{(j)}_{(2)}\right),\nonumber \\
&{\rm with} ~~ \mathbf{O}^{(j)} = \widetilde {\mathbf{F}}^{(j)} + \mathcal{C}_{spe}^{(j)}\left(\widetilde {\mathbf{F}}^{(j)}\right) \times \mathcal{M}\left(\widetilde{\mathbf{L}}^{(j)}_{(1)}\right).
    \label{con:eqution5}
\end{align}

Training such a masked Template-Net will lead to the posterior distribution $\mathcal{P}(\mathcal{G}|\mathbb{T})$. In the next section, we discuss the inference process of the proposed model and model epistemic uncertainty.

\subsection{Model Inference \& Epistemic Uncertainty}
\label{Sec:Inference}
As aforementioned, given an input LR-HS image, the proposed PDE-Net predicts the distribution of an HR-HS image, i.e.,  $\mathcal{P}(\mathbf{Y}|\mathbf{X},\mathbb{T})$. Thus,  we have to obtain its expectation to objectively compare reconstruction results. Specifically, we adopt the Monte Carlo (MC) sampling method to randomly sample $N$ models from $\mathcal{P}(\mathcal{G}|\mathbb{T})$, which output reconstructed HR-HS images denoted as $\widehat{\mathbf{Y}}_1,~\widehat{\mathbf{Y}}_2,~\cdots,~\widehat{\mathbf{Y}}_N\in\mathbb{R}^{B\times HW}$, and then calculate $\widehat{\mathbf{Y}}=\frac{1}{N}\sum_{n=1}^N\widehat{\mathbf{Y}}_n$.
Note that thanks to the parallelism of deep neural networks, we could realize the MC sampling efficiently via a batched inference manner, where we just feed $N$ copies of an input LR-HS image as a mini-batch and average the super-resolved HR-HS images in batch-wise. See Fig. \ref{fig:sampling} for the effect of the hyperparameter $N$ on quantitative reconstruction quality.

Based on the probabilistic characteristic of our PDE-Net, we can figure out the uncertainty of reconstruction by  calculating the probability of expectation.
To measure the variation of the reconstructed $\widehat{\mathbf{Y}}$, we first discretize the continuous space of the network output in the range of $[0,1]$ with an interval of $\frac{1}{255}$. Then we define the epistemic uncertainty of a pixel as
 \begin{align}
     &\mathcal{S}\left( \widehat{y} \right) = \sum_{n=1}^{N}{ \mathbf{I}_n } / {N} \times 100\%, \nonumber \\
     &{\rm with} ~~ \mathbf{I}_n = \left\{
     \begin{aligned}
        1 ~~ & {\rm if} ~\mathcal{Z}(\widehat{y}_n) \neq \mathcal{Z}(\widehat{y}), \\
        0 ~~ & {\rm otherwise},
    \end{aligned}
    \right.
 \label{eq:Reconstruction}
 \end{align}
where $\widehat{y}_n$ and $\widehat{y}$ are typical pixels of $\widehat{\mathbf{Y}}_n$ and $\widehat{\mathbf{Y}}$, respectively, and $\mathcal{Z}(\widehat{y})=\mathsf{round}(\widehat{y}\times 255)/255$  is the discretization function with $\mathsf{round}(\cdot)$ being the rounding operation.
Note that we do not require the ground-turth pixel value during the calculation of the epistemic uncertainty. Thus, we could measure the model epistemic uncertainty during both training and testing phases.

\subsection{Loss Function}
\label{ssec:lossfunction}
Following previous single-image and HS image super-resolution works \cite{Kim2016Accurate,Lim2017Enhanced,Zhang2018Image,Dai2019Second,Li2020Mixed,Jiang2020Learning}, we train PDE-Net by minimizing the $\ell_1$ distance between the $\mathbf{\widehat{Y}}$ and $\mathbf{Y}$:
\begin{equation}
\label{equ:l1loss}
\mathcal{L}_1(\mathbf{\widehat{Y}}, \mathbf{Y}) = \frac{1}{B\times HW}\left\|\mathbf{\widehat{Y}}- \mathbf{Y} \right\|_1.
\end{equation}
Besides, we also promote $\mathcal{D}(\mathbf{\widehat{Y}})$ to be close to $\mathbf{X}$ to regularize $\mathbf{\widehat{Y}}$, i.e.,
\begin{equation}
\label{equ:l2loss}
\mathcal{L}_2(\mathbf{\widehat{Y}}, \mathbf{X}) = \frac{1}{B\times hw}\left\|\mathcal{D}(\mathbf{\widehat{Y}})- \mathbf{X} \right\|_F^2,
\end{equation}
where $\|\cdot\|_F$ is the Frobenious norm of a matrix.
Thus, the overall loss function for training our PDE-Net is written as
\begin{equation}
\label{equ:loverall}
\mathcal{L}(\mathbf{\widehat{Y}}, \mathbf{Y}, \mathbf{X}) = \mathcal{L}_1(\mathbf{\widehat{Y}}, \mathbf{Y}) + \lambda\mathcal{L}_2(\mathbf{\widehat{Y}}, \mathbf{X}),
\end{equation}
where the hyper-parameter $\lambda$ is to balance these two terms, which is empirically set to 1.

\section{Experiments}
\label{sec:experiments}

\subsection{Experiment Settings}

\subsubsection{Datasets}
We employed 3 common HS image datasets to evaluate the performance of our PDE-Net, i.e.,  CAVE\footnote{http://www.cs.columbia.edu/CAVE/databases/} \cite{Yasuma2010CAVE}, Harvard\footnote{http://vision.seas.harvard.edu/hyperspec/} \cite{Chakrabarti2011Harvard},
and NCALM\footnote{http://www.grss-ieee.org/community/technical-committees/data-fusion/2018-ieee-grss-data-fusion-contest/}  \cite{xu2019advanced},
whose details are listed as follows.
\begin{itemize}
  \item The CAVE dataset contains 32 HS images of spatial dimensions $512\times512$ and spectral dimension 31, which were collected by a generalized assorted pixel camera ranging from 400 to 700 nm.
  We randomly selected 20 HS images for training, and the remaining 12 HS images for testing.
  \item The Harvard dataset consists of 50 HS images of  spatial dimensions $1040\times1392$ and spectral dimension 31, which were gathered by a Nuance FX, CRI Inc. camera covering the wavelength range from 420  to 720 nm.  We randomly selected 40 HS images as the training set , and the rest as the testing set.
  \item The NCALM dataset used for the IEEE GRSS Data Fusion Contest only contains one HS image of spatial dimensions $1202\times4172$, which covers a 380-1050 nm spectral range with 48 bands. For this image, we cropped four left regions of $512\times512$ spatial dimensions for testing and the rest for training.
\end{itemize}

During training, we cropped overlapped patches of spatial dimensions $128\times128$, and utilized rotation and flipping for data augmentation. Following previous works \cite{Jiang2020Learning,Li2020Mixed,Li2021Exploring}, we used the bicubic down-sampling method to generate LR-HS images.

\subsubsection{Implementation details} We implemented the proposed method with PyTorch, where the ADAM optimizer \cite{kingma2014adam} with the exponential decay rates $\beta_1=0.9$ and $\beta_2=0.999$ was utilized. We initialized the learning rate as $5\times10^{-4}$, which was halved every 25 epochs.
We set the batch size to 4 for all the three datasets. The total training process contained 50 warm-ups and 100 training epochs. During the warm-up phase, we set all elements of  $\{\mathbf{K}^{(j)},~\mathbf{L}^{(j)}\}_{j=1}^J$ to 1 for warming up the Template-Net. To increase the flexibility of our model, we finely defined the probability space in channel-wise, i.e., assigning different probabilities to each channel (convolutional kernel).

\subsubsection{Evaluation metrics} Following previous works
 \cite{Li2020Mixed}, \cite{Li2021Exploring}, we adopted three widely-used metrics to evaluate the quality of reconstructed HR-HS images quantitatively, i.e.,
mean peak signal-to-noise ratio (MPSNR), mean structure similarity (MSSIM) \cite{Zhou2002SSIM}, and spectral angle mapper (SAM) \cite{Yuhas1992Discrimination}.
For MPSNR and MSSIM, the larger, the better. For SAM, the smaller, the better. See \cite{zhu2021deep} for more details about the definitions of these metrics.

\subsection{Comparison with State-of-the-Art Methods}

\begin{table}[t]
\caption{Quantitative Comparisons Of Different Methods Over The CAVE Dataset. The best results of all methods and the best results of existing methods are highlighted in bold and underline, respectively.``$\uparrow$" (resp. ``$\downarrow$") means the larger (resp. smaller), the better.}
\vspace{-0.2cm}
\centering
\label{tab:caveresults}
\begin{tabu}{c|c|c|c|c|c}
\tabucline[1pt]{*}
Methods                                  &Scale  &\#Params &MPSNR$\uparrow$  &MSSIM$\uparrow$  &SAM$\downarrow$       \\  \hline\hline  
BI                                       &4    &-        &36.533    &0.9479   &4.230        \\
3DFCNN\cite{Mei2017Hyperspectral}        &4    &0.04M    &38.061    &0.9565   &3.912        \\
3DGAN\cite{Li2020Hyperspectral}          &4    &0.59M    &39.947    &0.9645   &3.702        \\
SSPSR\cite{Jiang2020Learning}            &4    &26.08M   &40.104    &0.9645   &3.623        \\
MCNet\cite{Li2020Mixed}                  &4    &2.17M    &40.658    &0.9662   &3.499        \\
ERCSR\cite{Li2021Exploring}              &4    &1.59M    &\underline{40.701}  &\underline{0.9662}   &\underline{3.491}        \\ \hline
Template-Net    &4    &2.29M    &40.911  &0.9666   &3.514        \\
PDE-Net         &4    &2.30M    &\textbf{41.236}    &\textbf{0.9672}   &\textbf{3.455}    \\ \hline\hline
BI                                       &8    &-        &32.283    &0.8993   &5.412        \\
3DFCNN\cite{Mei2017Hyperspectral}        &8    &0.04M    &33.194    &0.9131   &5.019        \\
3DGAN\cite{Li2020Hyperspectral}          &8    &0.66M    &34.930    &0.9293   &4.888        \\
SSPSR\cite{Jiang2020Learning}            &8    &28.44M   &34.992    &0.9273   &4.680        \\
MCNet\cite{Li2020Mixed}                  &8    &2.96M    &35.518    &0.9328   &4.519        \\
ERCSR\cite{Li2021Exploring}              &8    &2.38M    &\underline{35.519}  &\underline{0.9338}   &\underline{4.498}        \\ \hline
Template-Net      &8    &2.32M    &35.781 &0.9341   &4.442        \\
PDE-Net           &8    &2.33M   &\textbf{36.021}  &\textbf{0.9363}   &\textbf{4.312}        \\
\tabucline[1pt]{*}
\end{tabu}
\end{table}

\begin{table}[t]
\caption{Quantitative Comparisons Of Different Methods Over The Harvard Dataset. The best results of all methods and the best results of existing methods are highlighted in bold and underline, respectively.``$\uparrow$" (resp. ``$\downarrow$") means the larger (resp. smaller), the better.} \vspace{-0.2cm}
\centering
\label{tab:harvardresults}
\begin{tabu}{c|c|c|c|c|c}
\tabucline[1pt]{*}
Methods                                  &Scale  &\#Params  &MPSNR$\uparrow$  &MSSIM$\uparrow$  &SAM$\downarrow$  \\  \hline\hline  
BI                                       &4    &-          &37.255    &0.8977   &2.574     \\
3DFCNN\cite{Mei2017Hyperspectral}        &4    &0.04M      &38.110    &0.9101   &2.527     \\
3DGAN\cite{Li2020Hyperspectral}          &4    &0.59M      &38.781    &0.9189   &2.520     \\
SSPSR\cite{Jiang2020Learning}            &4    &26.08M     &39.397    &\underline{0.9287}   &\underline{2.433}     \\
MCNet\cite{Li2020Mixed}                  &4    &2.17M      &\underline{39.412}  &0.9268   &2.445     \\
ERCSR\cite{Li2021Exploring}              &4    &1.59M      &39.395    &0.9265   &2.440     \\ \hline
Template-Net                             &4    &2.29M      &39.595  &0.9295   &2.473      \\
PDE-Net                                  &4    &2.30M      &\textbf{40.021}     &\textbf{0.9346}   &\textbf{2.427}       \\ \hline\hline
BI                                       &8    &-          &33.597    &0.8129   &3.076       \\
3DFCNN\cite{Mei2017Hyperspectral}        &8    &0.04M      &34.155    &0.8251   &2.984       \\
3DGAN\cite{Li2020Hyperspectral}          &8    &0.66M      &34.799    &0.8321   &3.047       \\
SSPSR\cite{Jiang2020Learning}            &8    &28.44M     &35.094    &0.8410   &\textbf{2.871}     \\
MCNet\cite{Li2020Mixed}                  &8    &2.96M      &\underline{35.264}  &\underline{0.8414}   &2.937        \\
ERCSR\cite{Li2021Exploring}              &8    &2.38M      &35.207    &0.8402   &2.928       \\ \hline
Template-Net                             &8    &2.32M      &35.242    &0.8413   &2.983      \\
PDE-Net                                  &8    &2.33M     &\textbf{35.382}   &\textbf{0.8438}   &\underline{2.924}      \\
\tabucline[1pt]{*}
\end{tabu}
\end{table}

\begin{table}[!t]
\caption{Quantitative Comparisons Of Different Methods Over The NCALM Dataset. The best results of all methods and the best results of existing methods are highlighted in bold and underline, respectively.``$\uparrow$" (resp. ``$\downarrow$") means the larger (resp. smaller), the better.}\vspace{-0.2cm}
\centering
\label{tab:ieeecontestresults}
\begin{tabu}{c|c|c|c|c|c}
\tabucline[1pt]{*}
Methods                                  &Scale  &\#Params  &MPSNR$\uparrow$  &MSSIM$\uparrow$  &SAM$\downarrow$  \\  \hline\hline  
BI                                       &4    &-         &43.618   &0.9646   &2.504     \\
3DFCNN\cite{Mei2017Hyperspectral}        &4    &0.04M     &44.300   &0.9703   &2.390     \\
3DGAN\cite{Li2020Hyperspectral}          &4    &0.59M     &45.239   &0.9761   &2.267     \\
SSPSR\cite{Jiang2020Learning}            &4    &12.88M    &45.271   &0.9754   &2.221     \\
MCNet\cite{Li2020Mixed}                  &4    &2.17M     &45.578   &0.9764   &2.156     \\
ERCSR\cite{Li2021Exploring}              &4    &1.59M     &\underline{45.683}   &\underline{0.9768}   &\underline{2.132}     \\ \hline
Template-Net                             &4    &2.29M     &45.920 &0.9780 &2.155   \\
PDE-Net                                  &4    &2.30M     &\textbf{46.533}    &\textbf{0.9810}    &\textbf{1.927}       \\ \hline\hline
BI                                       &8    &-         &38.699    &0.9079   &4.530    \\
3DFCNN\cite{Mei2017Hyperspectral}        &8    &0.04M     &39.128    &0.9142   &4.409    \\
3DGAN\cite{Li2020Hyperspectral}          &8    &0.66M     &39.527    &0.9190   &4.272    \\
SSPSR\cite{Jiang2020Learning}            &8    &15.23M    &39.799    &0.9221   &4.150    \\
MCNet\cite{Li2020Mixed}                  &8    &2.96M     &39.809    &0.9217   &4.153    \\
ERCSR\cite{Li2021Exploring}              &8    &2.38M     &\underline{39.999}    &\underline{0.9233}  &\underline{4.103}     \\ \hline
Template-Net                             &8    &2.32M    &40.007  &0.9225    &4.244        \\
PDE-Net                                  &8    &2.33M     &\textbf{40.286}   &\textbf{0.9265}   &\textbf{3.976}  \\
\tabucline[1pt]{*}
\end{tabu}
\end{table}

\begin{table}[!t]
\scriptsize
\caption{Results of the ablation study towards the computational efficiency over the CAVE dataset. $N$ refers to the MC sampling times. } \vspace{-0.2cm}
\centering
\label{tab:flops-results}
\setlength{\tabcolsep}{0.5mm}{
\begin{tabu}{c|c|c|c|c|c|c}  \tabucline[1pt]{*}

Methods &Scale
&Inference time &\#FLOPs &Scale  &Inference time &\#FLOPs   \\ \hline

3DFCNN\cite{Mei2017Hyperspectral}  &4  &0.197s  &0.321T      &8   &0.195s   &0.321T    \\
3DGAN\cite{Li2020Hyperspectral}    &4  &0.382s  &1.300T      &8   &0.337s   &1.233T    \\
SSPSR\cite{Jiang2020Learning}      &4  &0.429s  &3.029T      &8   &0.251s   &1.818T    \\
MCNet\cite{Li2020Mixed}            &4  &0.578s  &4.489T      &8   &0.327s   &10.220T   \\
ERCSR\cite{Li2021Exploring}        &4  &0.430s  &4.463T      &8   &0.266s   &10.429T   \\ \hline
PDE-Net ($N=1$)           &4  &0.641s  &1.275T &8  &0.189s   &0.604T   \\
PDE-Net ($N=5$)           &4  &0.704s  &6.375T &8  &0.258s   &3.020T  \\
\tabucline[1pt]{*}
\end{tabu}}
\end{table}

\begin{figure*}[!t]
\centering
\includegraphics[width=0.9\linewidth]{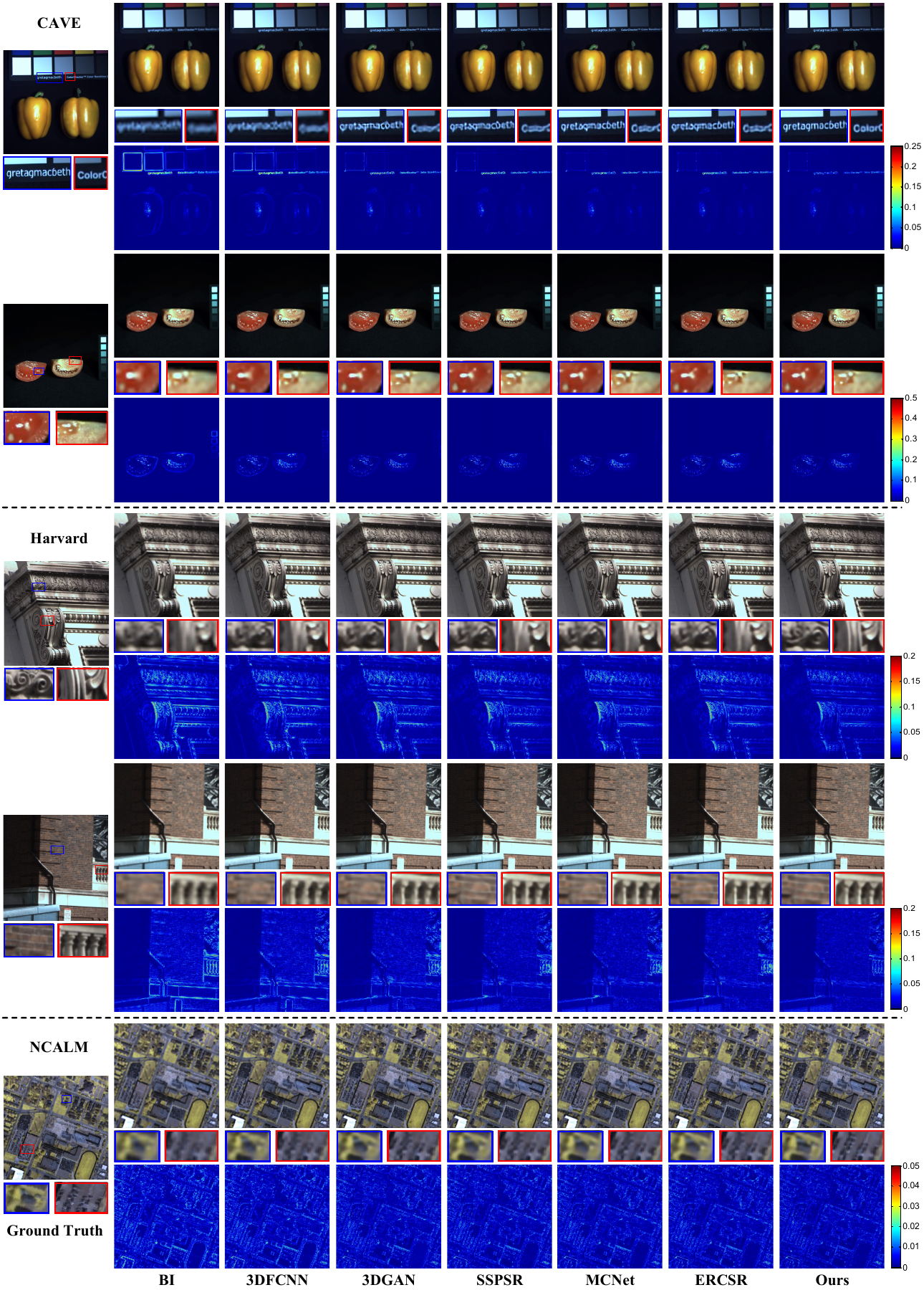}\vspace{-0.2cm}
\caption{Visual comparisons of different methods with $\alpha=4$.
For ease of comparison, we visualized the reconstructed HS images in the form of RGB images, which were generated via employing the commonly-used spectral response function of Nikon-D700
\cite{jiang2013space}.
}
\label{fig:Harvard-errormap}
\end{figure*}

\begin{figure*}[t]
\centering
\includegraphics[width=1\linewidth]{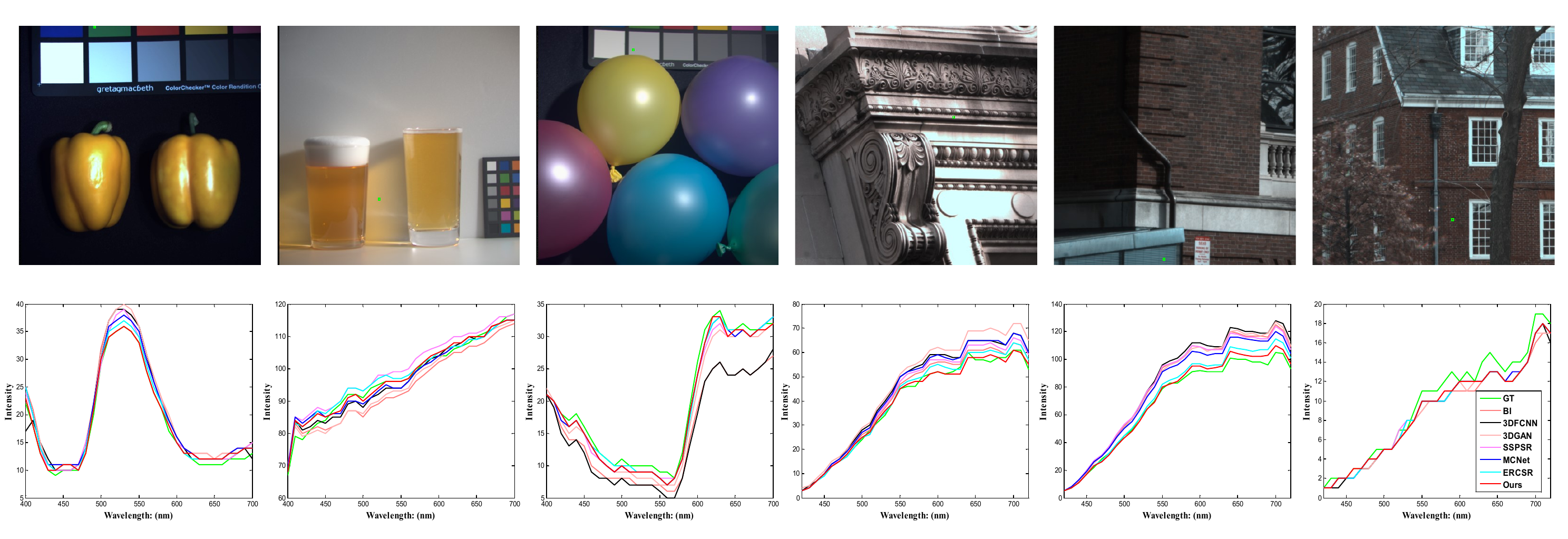}\vspace{-0.3cm}
\caption{Visual comparison of the spectral signatures of pixels reconstructed by different methods.
The positions of the corresponding pixels are marked by the green dot in RGB images. The spectral signatures by our PDE-Net are much closer to the ground-truth ones than the other compared methods, especially on the $1^{st}$ and $4^{th}$ columns.}
\label{fig:spectral-intensity}
\end{figure*}

\begin{figure}[!t]
    \centering
    \includegraphics[width=1.0\linewidth]{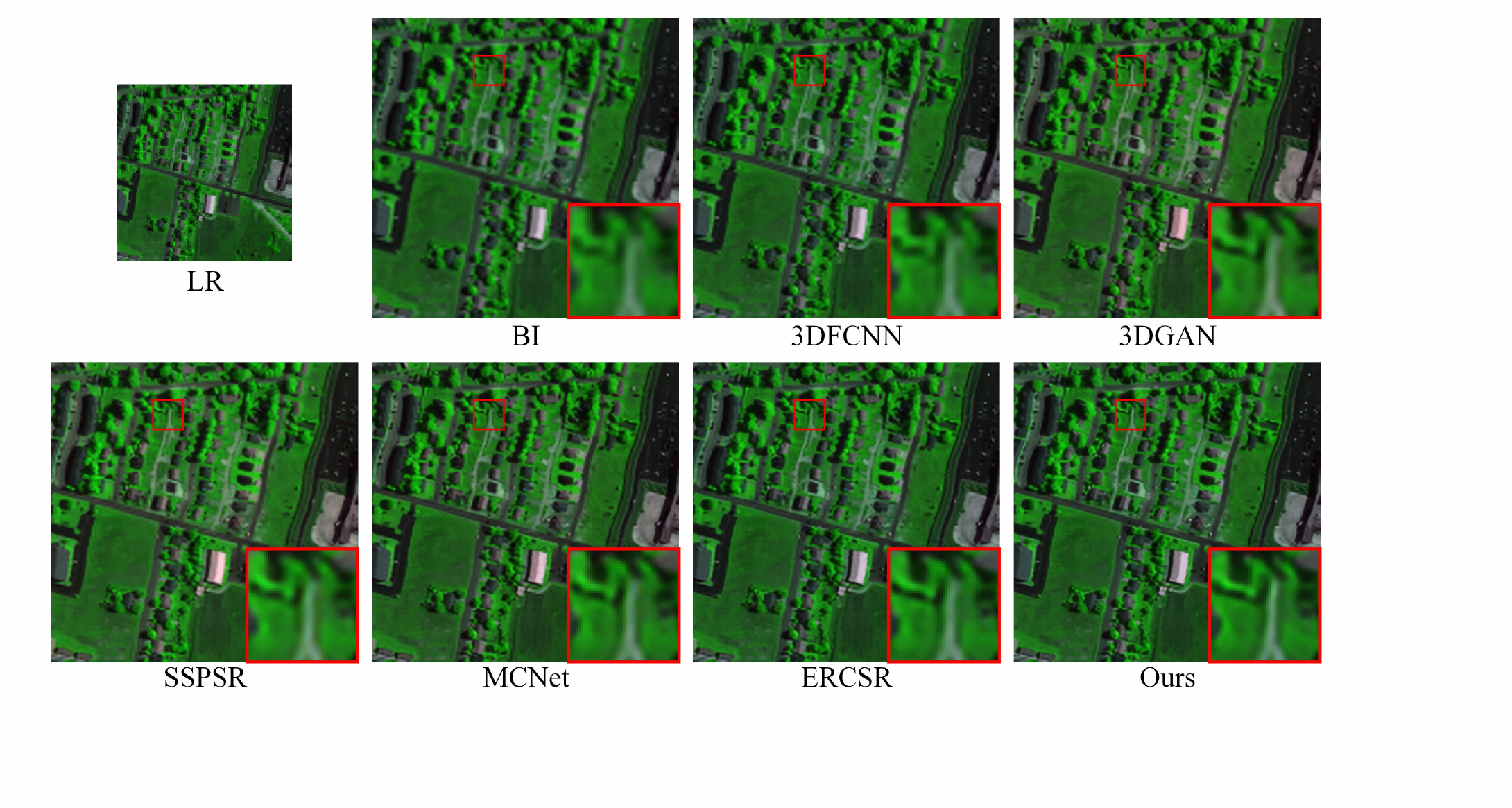}\vspace{-0.3cm}
    \caption{Visual comparison of different methods on the HS image from the Urban dataset ($\alpha=4$).}
    \label{fig:realtest}
\end{figure}

\begin{figure*}[!t]
\centering
\includegraphics[width=1\linewidth]{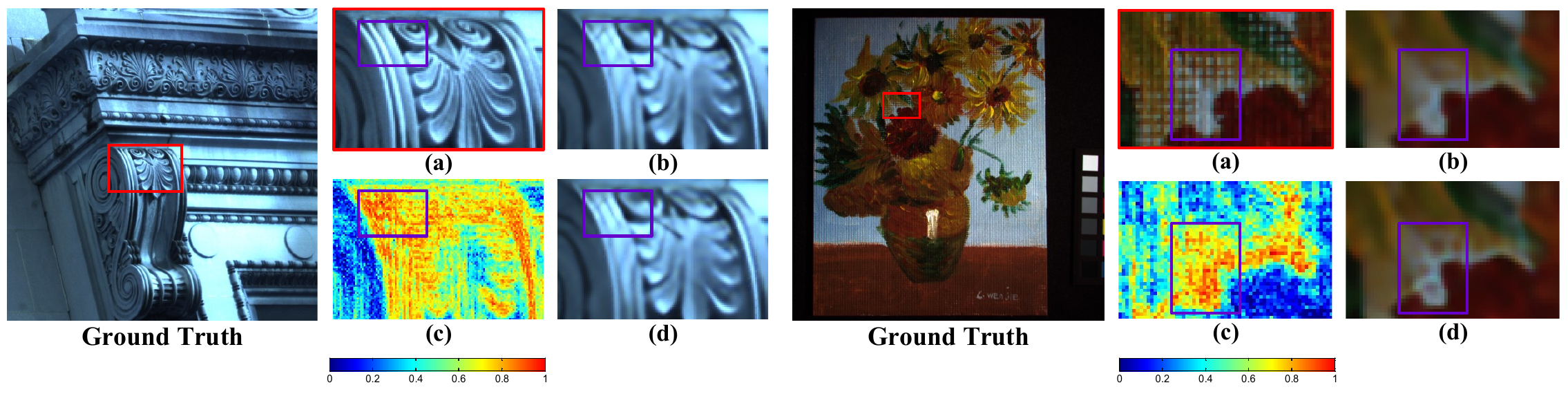} \vspace{-0.3cm}
\caption{Visual illustration of the uncertainty estimation.
(a) Zoomed-in patch in the red frame of the ground-truth HS images; (b) and (d) reconstructed HS images via two MC sampling;
and (c): uncertainty maps. It could be observed that higher uncertainties correspond to larger variations of textures and higher risks of reconstruction errors.}
\label{fig:uncertainty}
\end{figure*}

We compared the proposed PDE-Net with 5 state-of-the-art deep learning-based methods, i.e., 3DFCNN \cite{Mei2017Hyperspectral}, 3DGAN \cite{Li2020Hyperspectral}, SSPSR \cite{Jiang2020Learning}, MCNet \cite{Li2020Mixed}, and ERCSR \cite{Li2021Exploring}. We also provided the results of bi-cubic interpolation (BI) as a baseline. For a fair comparison,
we retrained all the compared methods with the same training data as ours by using the codes released by the authors with suggested settings. Besides, we applied the same data pre-processing to all methods.

\begin{figure}[!t]
\centering
\includegraphics[width=0.9\linewidth]{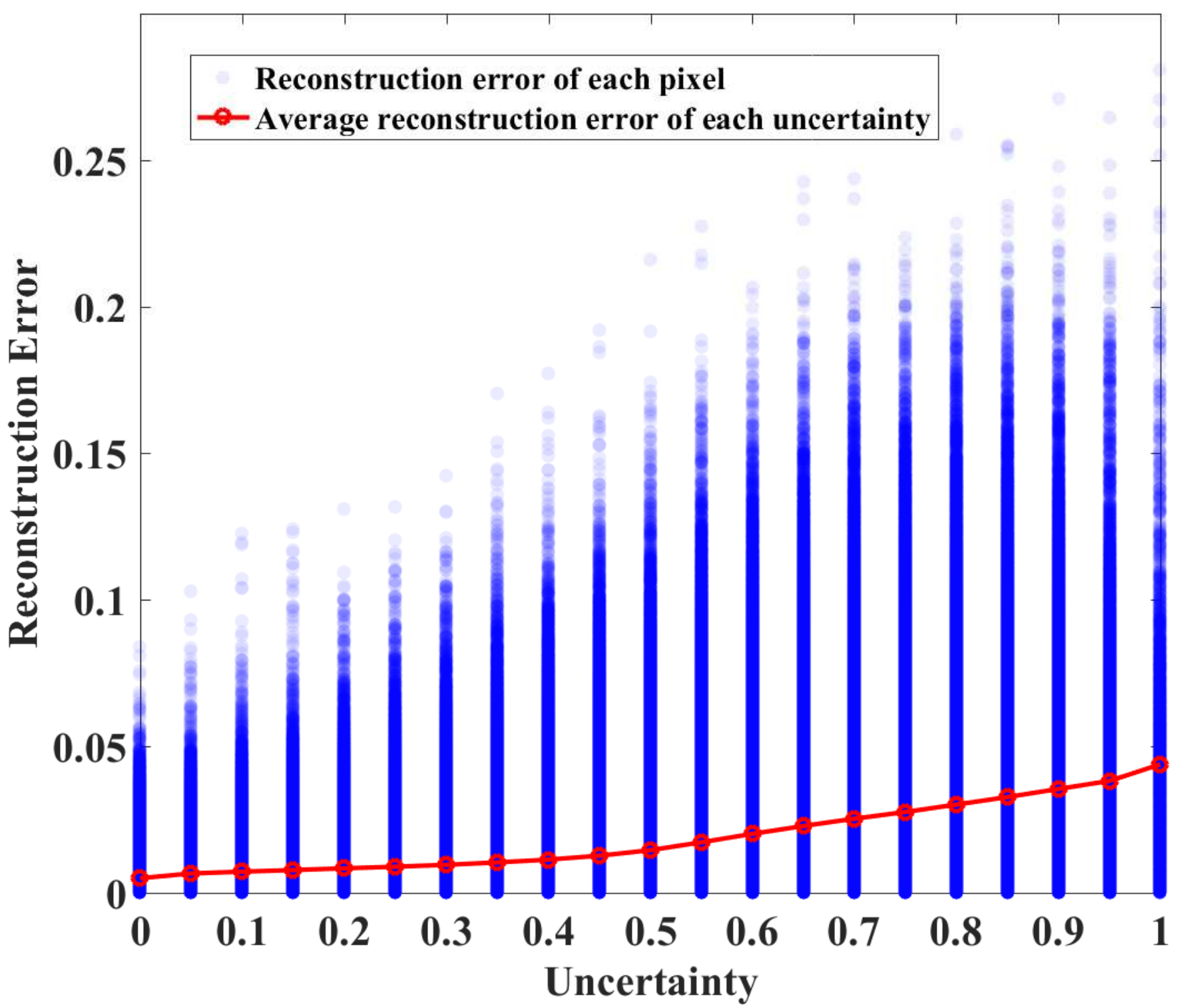}\vspace{-0.3cm}
\caption{Visualization of the relationship between the uncertainty and reconstruction error of pixels.
It can be observed that the average reconstruction error is approximately proportional to the pixel uncertainty.}
\label{fig:relationship_u_e}
\end{figure}

\begin{figure*}[!t]
\centering
\includegraphics[width=0.95\linewidth]{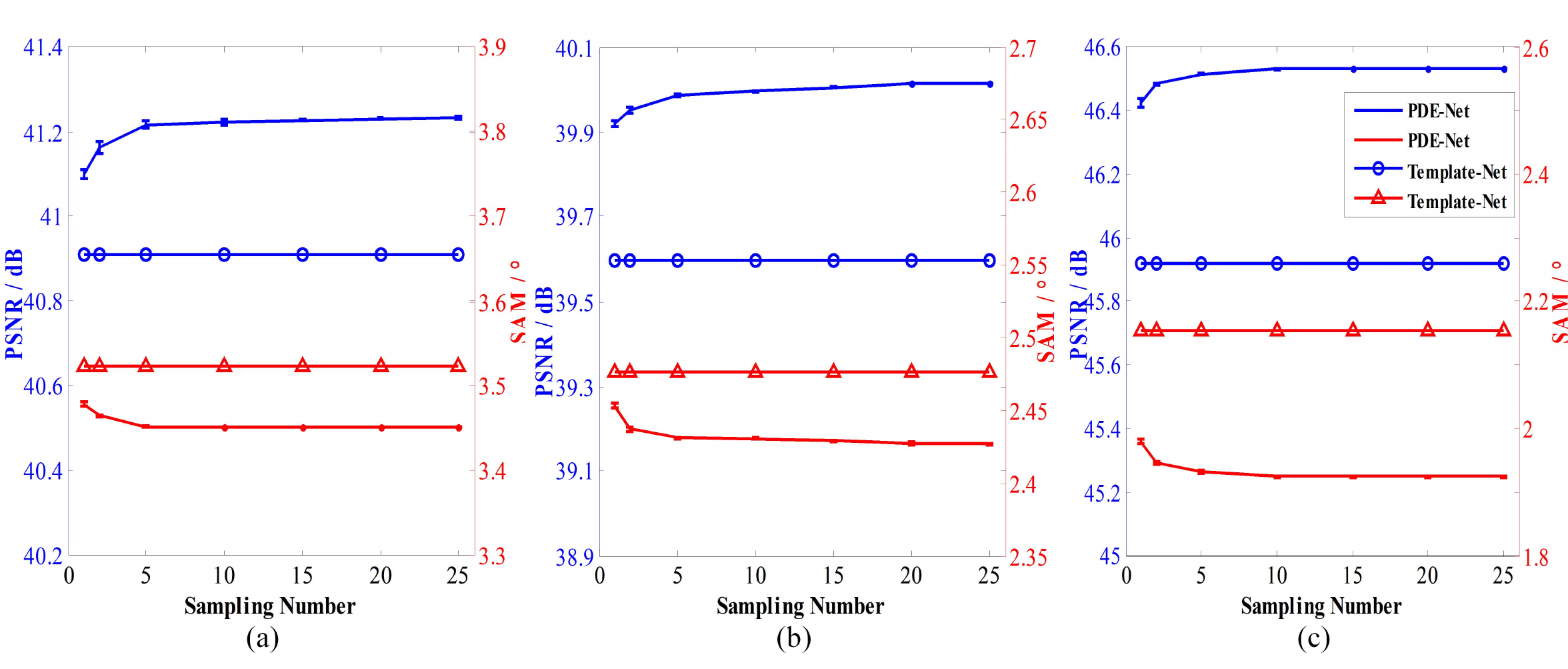} \vspace{-0.3cm}
\caption{Illustration of the performance of our PDE-Net and Template-Net with different MC sampling times on three datasets ($\alpha=4$), i.e., (a) CAVE, (b) Harvard, and (c) NCALM.
}
\label{fig:sampling}
\end{figure*}

Tables \ref{tab:caveresults}, \ref{tab:harvardresults}, and \ref{tab:ieeecontestresults} show the quantitative results of different methods on the three datasets, where it can be observed that
\begin{itemize}
   \item our PDE-Net consistently achieves the best performance in terms of all the three metrics on  all the three datasets when $\alpha=4$ and $8$, except the SAM value on the Harvard dataset for the $8\times$ super-resolution. Especially, our PDE-Net improves the MPSNR of the best existing methods by $0.53$ dB, $0.61$ dB, and $0.85$ dB (resp. $0.50$ dB, $0.12$ dB, and $0.29$ dB) over the CAVE, Harvard, and NCALM, respectively, when  $\alpha=4$ (resp. 8). Moreover, the superiority of SSPSR \cite{Jiang2020Learning} over our PDE-Net in terms of SAM under the $8\times$ super-resolution may
  benefit from the huge number of network parameters and the adopted spectral attention mechanism;

  \item the proposed Template-Net also obtains better reconstruction quality than most of the compared methods, demonstrating the superiority of our source-consistent HS images reconstruction framework to some extent;

  \item our PDE-Net further improves the Template-Net on the three datasets under all scenarios, validating the effectiveness and advantage of our posterior distribution-based HS embedding method; and

  \item for ERCSR \cite{Li2021Exploring} that always achieves the best or second-best performance among the compared methods, although it has a smaller number of network parameters than our PDE-Net, increasing its number of parameters cannot bring obvious performance improvement or even worsens performance \cite{Li2021Exploring}, due to the network architecture limitation. Besides, as listed in Table~\ref{tab:stagesresults} our PDE-Net with a comparable number of parameters to ERCSR still achieves better performance than ERCSR \cite{Li2021Exploring}.
\end{itemize}

Besides, Fig. \ref{fig:Harvard-errormap} visually compares the results by different methods, where we can observe that most high-frequency details are lost in the super-resolved images by the compared methods. By contrast,  our PDE-Net produces results with sharper textures closer to the
ground truth ones, which further demonstrates its advantage. In addition, Fig. \ref{fig:spectral-intensity} illustrates the spectral signatures of some pixels of reconstructed HR-HS images by different methods, where it can be seen that the shapes of the spectral signatures of all methods are generally consistent with those of the ground-truth ones. Moreover, the spectral signatures by our PDE-Net are closer to the ground-truth ones than the other methods, demonstrating the advantage of our method.

\begin{figure}[!t]
\centering
\includegraphics[width=0.9\linewidth]{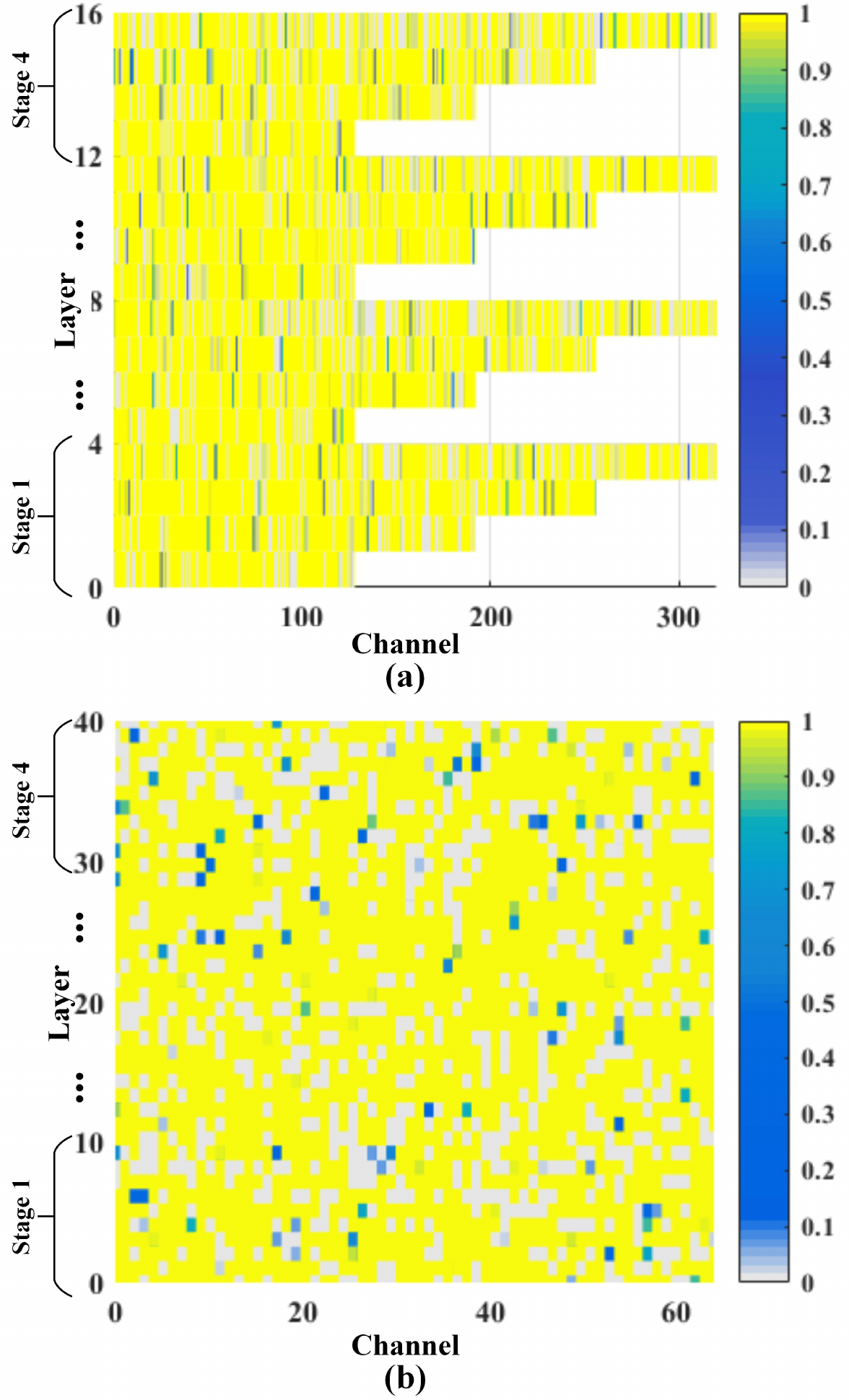}\vspace{-0.3cm}
\caption{Visualization of network-level and layer-level probability space corresponding to the channels and layers of  PDE-Net on the CAVE dataset for $4\times$ SR. (a) and (b) denote the network-level and layer-level distribution, respectively.
}
\label{fig:probability_distribution}
\end{figure}

To demonstrate the robustness and generalization of our PDE-Net in practice,  we also conducted the experiment in a real scenario, in which the input LR-HS is directly acquired by a typical sensor but not simulated by spatially downsampling the corresponding HR-HS image. Specifically, we utilized an HS image of spatial dimensions $307\times307$ and spectral dimension 210 ranging from 400 to 2500 nm from the Urban\footnote{https://rslab.ut.ac.ir/data} dataset collected by the HYDICE hyperspectral system.
Due to the limitation of computing resources, we only selected a region of size $128 \times 128$ from the HS image for testing.
Fig. \ref{fig:realtest} visually compares the results  of different methods trained with the NCALM dataset, where it can be seen that the super-resolved image by our method shows clearer and sharper textures, demonstrating the advantage of our method. Note that the corresponding HR ground-truth HS image is not unknown, making it impossible to quantitatively compare different methods here.

Finally, we compared the computational efficiency of different methods measured with the inference time and the number of floating point of operations (\#FLOPs) in Table \ref{tab:flops-results}.
It can be seen that PDE-Net consumes less inference time and much fewer \#FLOPs than most compared methods when performing MC sampling only once ($N=1$). Although \#FLOPs grows linearly with the number of MC sampling (i.e., the value of $N$) increasing, we want to note that the MC-sampling process could be realized in a parallel manner as mentioned in Section \ref{Sec:Inference}, and thus with more GPU nodes, the inference time of $N$ MC sampling could be comparable to that of 1 MC sampling.
Besides, as illustrated in Fig. \ref{fig:sampling}, the reconstruction quality increases relatively rapidly in the first 5 MC sampling but marginally when performing more MC sampling. Thus, in practice, one can perform MC sampling 5 times at most to save computational cost only with slight reconstruction quality sacrifice.

\subsection{Ablation Study}

\subsubsection{The number of stages}
To explore how  the number of stages involved in our PDE-Net affects performance, we evaluated the PDE-Net with various numbers of stages, i.e.,  $T=2,~3,~4,~5,~6$ and $7$.
From Table \ref{tab:stagesresults}, we can see that increasing the number of stages appropriately is able to improve the performance of both PDE-Net and Template-Net,
demonstrating the rationality of the iterative refinement strategy on our source-consistent reconstruction framework.
Especially, the PDE-Net is consistently better than Template-Net under all scenarios, which further indicates the effectiveness of our posterior distribution-based HS embedding method.
Observing that when $T>4$, the performance of Template-Net is almost stable, and PDE-Net improves very slightly,
we set the number of stages of our PDE-Net to 4 in all the remaining experiments of this paper.

\subsubsection{The $\mathcal{L}_2$ loss}
Table \ref{tab:wol2loss} lists the reconstruction quality of our PDE-Net with and without the $\mathcal{L}_2$ loss during training, where it can be concluded that the $\mathcal{L}_2$ loss makes contributions to the reconstruction process of our PDE-Net. The reason is that employing $\mathcal{L}_2$ loss can not only regularize the reconstructed HR-HS image, but also guarantee the residual between the pseudo-LR-HS image and the input LR-HS image can be minimized progressively.

\begin{table}[!t]
\caption{Results of the ablation study towards the number of stages ($T$) over the CAVE dataset ($\alpha=4$).
}\vspace{-0.2cm}
\centering
\label{tab:stagesresults}
\begin{tabu}{c|c|c|c|c|c}  
\tabucline[1pt]{*}
Stages              &Methods       &\#Params  &MPSNR$\uparrow$  &MSSIM$\uparrow$  &SAM$\downarrow$   \\  \hline  
\multirow{2}{*}{2}  &Template-Net   &1.147M  &40.785  &0.9663   &3.556   \\
                    &PDE-Net        &1.151M  &40.997  &0.9668   &3.477   \\ \hline
\multirow{2}{*}{3}  &Template-Net   &1.721M  &40.879  &0.9667   &3.520   \\
                    &PDE-Net        &1.726M  &41.145  &0.9671   &3.457   \\ \hline
\multirow{2}{*}{4}  &Template-Net   &2.295M  &40.911  &0.9666   &3.514   \\
                    &PDE-Net        &2.301M  &41.236  &0.9672   &3.455   \\ \hline
\multirow{2}{*}{5}  &Template-Net   &2.868M  &41.047  &0.9666   &3.509   \\
                    &PDE-Net        &2.877M  &41.241  &0.9672   &3.437   \\ \hline
\multirow{2}{*}{6}  &Template-Net   &3.442M  &41.027  &0.9664   &3.497   \\
                    &PDE-Net        &3.452M  &41.257  &0.9670   &3.439   \\ \hline
\multirow{2}{*}{7}  &Template-Net   &4.016M  &41.049  &0.9667   &3.495   \\
                    &PDE-Net        &4.028M  &41.307  &0.9674   &3.434   \\
\tabucline[1pt]{*}
\end{tabu}
\end{table}

\begin{table}[!t]
\caption{Results of the ablation study towards the $\mathcal{L}_2$ loss over the CAVE dataset ($\alpha=4$).} \vspace{-0.2cm}
\centering
\label{tab:wol2loss}
\begin{tabu}{c|ccc}
\tabucline[1pt]{*}
Methods        &MPSNR$\uparrow$  &MSSIM$\uparrow$  &SAM$\downarrow$   \\  \hline  
PDE-Net  w/o $\mathcal{L}_2$   &41.083  &0.9670  &3.467   \\
PDE-Net                        &41.236  &0.9672  &3.455  \\
\tabucline[1pt]{*}
\end{tabu}
\end{table}

\begin{table}[!t]
\caption{Comparison of the proposed posterior distribution-based  and NAS-based embedding schemes on the CAVE dataset ($\alpha=4$).} \vspace{-0.2cm}
\centering
\label{tab:nasvspde}
\begin{tabu}{c|ccc}
\tabucline[1pt]{*}
Methods        &MPSNR$\uparrow$  &MSSIM$\uparrow$  &SAM$\downarrow$   \\  \hline  
NAS-based            &41.085    &0.9668    &3.490   \\
PDE-Net        &41.236    &0.9672    &3.455   \\
\tabucline[1pt]{*}
\end{tabu}
\end{table}

\subsubsection{Illustration of the epistemic uncertainty}
As shown in Figs. \ref{fig:uncertainty} and \ref{fig:relationship_u_e}, as expected,
the high uncertainty always occurs in the regions with highly-volatile textures and large reconstruction errors.
Therefore, such epistemic uncertainty maps could help us to figure out the regions that are hard to handle,
so that additional efforts or more advanced super-resolution techniques can be considered to improve these regions. Moreover, the predicted uncertainty may also give the confidence of network outputs in other HS image-based high-level applications, such as, HS image classification (assigning pixel-wise object categories to HS images) \cite{mou2017deep,hong2020graph} and object detection/tracking \cite{liang2018material,zhang2018salient}\cite{uzkent2017aerial,tochon2017object}.

\subsubsection{MC sampling} We validated how the MC sampling times affects the performance of our PDE-Net.
Specifically, we calculated the mean value and standard deviation of MPSNRs/SAMs obtained via multiple MC sampling.
As shown in  Fig. \ref{fig:sampling}, it can be observed that the PDE-Net consistently outperforms the Template-Net over all the three datasets. As the number of MC sampling gradually rising up, the average value of samples is gradually approaching the expectation of the distribution.
Thus, the performance of PDE-Net gradually rises and finally achieves stable with the MC sampling times increasing.

\subsubsection{Visualization of the learned posterior distribution}
To have an intuitive understanding of our HS embedding architecture adaptively learned from the probabilistic perspective, we visualized the learned network-level and layer-level distributions in Fig. \ref{fig:probability_distribution}, where we can observe that the layer-level distribution is generally more complex than the network-level distribution, which is credited to the need of spatial-spectral diversities of local feature embedding.

\subsubsection{Posterior Distribution-based  vs. Network Architecture Search (NAS)-based embedding schemes}
NAS-based schemes learn the network topology via maximum a posterior distribution \cite{Liu2019DARTS}, resulting in a determined network architecture.
Although such an optimization scheme may produce the most possible model among the whole feasible set, compared to the proposed posterior distribution-based embedding, it discards a great number of plausible cases, which may also fit training samples well and contribute to performance improvement.
To quantitatively compare these two embedding schemes, we constructed an NAS-based framework, in which with the same training data as ours and well-tuned hyperparameters, we trained our framework with the NAS strategy, to optimize the topology of the set of feasible HS embedding events $\mathcal{G}$. As shown in Table \ref{tab:nasvspde}, our PDE-Net surpasses the NAS-based method in terms of all three metrics, demonstrating the advantage of our posterior distribution-based embedding scheme.

\section{Conclusion}
\label{sec:con}

We have proposed PDE-Net, a novel end-to-end learning-based framework for HS image super-resolution.
We built PDE-Net on the basis of the intrinsic degradation relationship between LR and HR-HS images,  thus making it physically-interpretable and compact.
More importantly, we formulated HS embedding, a core module contained in the PDE-Net,
from the probabilistic perspective to extract the high-dimensional spatial-spectral information efficiently and effectively.
By conducting extensive experiments on three common datasets, we demonstrated the significant superiority of our PDE-Net over state-of-the-art methods both quantitatively and qualitatively. Besides, we provided comprehensive ablation studies to have a better understanding of the proposed PDE-Net.

\ifCLASSOPTIONcaptionsoff
  \newpage
\fi

\balance
\bibliographystyle{IEEEtran}
\bibliography{HSISR}

\begin{thebibliography}{10}
\providecommand{\url}[1]{#1}
\csname url@samestyle\endcsname
\providecommand{\newblock}{\relax}
\providecommand{\bibinfo}[2]{#2}
\providecommand{\BIBentrySTDinterwordspacing}{\spaceskip=0pt\relax}
\providecommand{\BIBentryALTinterwordstretchfactor}{4}
\providecommand{\BIBentryALTinterwordspacing}{\spaceskip=\fontdimen2\font plus
\BIBentryALTinterwordstretchfactor\fontdimen3\font minus
  \fontdimen4\font\relax}
\providecommand{\BIBforeignlanguage}[2]{{%
\expandafter\ifx\csname l@#1\endcsname\relax
\typeout{** WARNING: IEEEtran.bst: No hyphenation pattern has been}%
\typeout{** loaded for the language `#1'. Using the pattern for}%
\typeout{** the default language instead.}%
\else
\language=\csname l@#1\endcsname
\fi
#2}}
\providecommand{\BIBdecl}{\relax}
\BIBdecl

\bibitem{park2015hyperspectral}
B.~Park and R.~Lu, \emph{Hyperspectral imaging technology in food and
  agriculture}.\hskip 1em plus 0.5em minus 0.4em\relax Springer, 2015.

\bibitem{lu2020recent}
B.~Lu, P.~D. Dao, J.~Liu, Y.~He, and J.~Shang, ``Recent advances of
  hyperspectral imaging technology and applications in agriculture,''
  \emph{Remote Sensing}, vol.~12, no.~16, p. 2659, 2020.

\bibitem{shimoni2019hypersectral}
M.~Shimoni, R.~Haelterman, and C.~Perneel, ``Hypersectral imaging for military
  and security applications: Combining myriad processing and sensing
  techniques,'' \emph{IEEE Geoscience and Remote Sensing Magazine}, vol.~7,
  no.~2, pp. 101--117, 2019.

\bibitem{jia2020status}
J.~Jia, Y.~Wang, J.~Chen, R.~Guo, R.~Shu, and J.~Wang, ``Status and application
  of advanced airborne hyperspectral imaging technology: A review,''
  \emph{Infrared Physics \& Technology}, vol. 104, p. 103115, 2020.

\bibitem{banerjee2020uav}
B.~P. Banerjee, S.~Raval, and P.~Cullen, ``Uav-hyperspectral imaging of
  spectrally complex environments,'' \emph{International Journal of Remote
  Sensing}, vol.~41, no.~11, pp. 4136--4159, 2020.

\bibitem{mishra2017close}
P.~Mishra, M.~S.~M. Asaari, A.~Herrero-Langreo, S.~Lohumi, B.~Diezma, and
  P.~Scheunders, ``Close range hyperspectral imaging of plants: A review,''
  \emph{Biosystems Engineering}, vol. 164, pp. 49--67, 2017.

\bibitem{Mei2017Hyperspectral}
S.~Mei, X.~Yuan, J.~Ji, Y.~Zhang, S.~Wan, and Q.~Du, ``Hyperspectral image
  spatial super-resolution via 3d full convolutional neural network,''
  \emph{Remote Sensing}, vol.~9, no.~11, p. 1139, 2017.

\bibitem{Li2020Hyperspectral}
J.~Li, R.~Cui, B.~Li, R.~Song, Y.~Li, Y.~Dai, and Q.~Du, ``Hyperspectral image
  super-resolution by band attention through adversarial learning,'' \emph{IEEE
  Transactions on Geoscience and Remote Sensing}, vol.~58, no.~6, pp.
  4304--4318, 2020.

\bibitem{Li2021Exploring}
Q.~Li, Q.~Wang, and X.~Li, ``Exploring the relationship between 2d/3d
  convolution for hyperspectral image super-resolution,'' \emph{IEEE
  Transactions on Geoscience and Remote Sensing}, pp. 1--11, 2021.

\bibitem{Li2020Mixed}
Q.~{Li}, Q.~{Wang}, and X.~{Li}, ``Mixed 2d/3d convolutional network for
  hyperspectral image super-resolution,'' \emph{Remote Sensing}, vol.~12,
  no.~10, p. 1660, 2020.

\bibitem{Jiang2020Learning}
J.~{Jiang}, H.~{Sun}, X.~{Liu}, and J.~{Ma}, ``Learning spatial-spectral prior
  for super-resolution of hyperspectral imagery,'' \emph{IEEE Transactions on
  Computational Imaging}, vol.~6, pp. 1082--1096, 2020.

\bibitem{Dong2016Hyperspectral}
W.~{Dong}, F.~{Fu}, G.~{Shi}, X.~{Cao}, J.~{Wu}, G.~{Li}, and X.~{Li},
  ``Hyperspectral image super-resolution via non-negative structured sparse
  representation,'' \emph{IEEE Transactions on Image Processing}, vol.~25,
  no.~5, pp. 2337--2352, 2016.

\bibitem{yi2018hyperspectral}
C.~Yi, Y.-Q. Zhao, and J.~C.-W. Chan, ``Hyperspectral image super-resolution
  based on spatial and spectral correlation fusion,'' \emph{IEEE Transactions
  on Geoscience and Remote Sensing}, vol.~56, no.~7, pp. 4165--4177, 2018.

\bibitem{Akhtar2015Bayesian}
N.~Akhtar, F.~Shafait, and A.~Mian, ``Bayesian sparse representation for
  hyperspectral image super resolution,'' in \emph{Proc. IEEE/CVF Conference on
  Computer Vision and Pattern Recognition}, 2015, pp. 3631--3640.

\bibitem{xu2019nonlocal}
Y.~Xu, Z.~Wu, J.~Chanussot, and Z.~Wei, ``Nonlocal patch tensor sparse
  representation for hyperspectral image super-resolution,'' \emph{IEEE
  Transactions on Image Processing}, vol.~28, no.~6, pp. 3034--3047, 2019.

\bibitem{Han2018Self}
X.~{Han}, B.~{Shi}, and Y.~{Zheng}, ``Self-similarity constrained sparse
  representation for hyperspectral image super-resolution,'' \emph{IEEE
  Transactions on Image Processing}, vol.~27, no.~11, pp. 5625--5637, 2018.

\bibitem{Dian2017Hyperspectral}
R.~Dian, L.~Fang, and S.~Li, ``Hyperspectral image super-resolution via
  non-local sparse tensor factorization,'' in \emph{Proc. IEEE/CVF Conference
  on Computer Vision and Pattern Recognition}, 2017, pp. 3862--3871.

\bibitem{Dian2019Learning}
R.~{Dian}, S.~{Li}, and L.~{Fang}, ``Learning a low tensor-train rank
  representation for hyperspectral image super-resolution,'' \emph{IEEE
  Transactions on Neural Networks and Learning Systems}, vol.~30, no.~9, pp.
  2672--2683, 2019.

\bibitem{dian2019hyperspectral}
R.~Dian and S.~Li, ``Hyperspectral image super-resolution via subspace-based
  low tensor multi-rank regularization,'' \emph{IEEE Transactions on Image
  Processing}, vol.~28, no.~10, pp. 5135--5146, 2019.

\bibitem{xue2021spatial}
J.~Xue, Y.-Q. Zhao, Y.~Bu, W.~Liao, J.~C.-W. Chan, and W.~Philips,
  ``Spatial-spectral structured sparse low-rank representation for
  hyperspectral image super-resolution,'' \emph{IEEE Transactions on Image
  Processing}, vol.~30, pp. 3084--3097, 2021.

\bibitem{vicinanza2014pansharpening}
M.~R. Vicinanza, R.~Restaino, G.~Vivone, M.~Dalla~Mura, and J.~Chanussot, ``A
  pansharpening method based on the sparse representation of injected
  details,'' \emph{IEEE Geoscience and Remote Sensing Letters}, vol.~12, no.~1,
  pp. 180--184, 2014.

\bibitem{fei2019convolutional}
R.~Fei, J.~Zhang, J.~Liu, F.~Du, P.~Chang, and J.~Hu, ``Convolutional sparse
  representation of injected details for pansharpening,'' \emph{IEEE Geoscience
  and Remote Sensing Letters}, vol.~16, no.~10, pp. 1595--1599, 2019.

\bibitem{Xie2019Multispectral}
Q.~{Xie}, M.~{Zhou}, Q.~{Zhao}, D.~{Meng}, W.~{Zuo}, and Z.~{Xu},
  ``Multispectral and hyperspectral image fusion by ms/hs fusion net,'' in
  \emph{Proc. IEEE/CVF Conference on Computer Vision and Pattern Recognition},
  2019, pp. 1585--1594.

\bibitem{yao2020cross}
J.~Yao, D.~Hong, J.~Chanussot, D.~Meng, X.~Zhu, and Z.~Xu, ``Cross-attention in
  coupled unmixing nets for unsupervised hyperspectral super-resolution,'' in
  \emph{Proc. European Conference on Computer Vision}.\hskip 1em plus 0.5em
  minus 0.4em\relax Springer, 2020, pp. 208--224.

\bibitem{qu2018unsupervised}
Y.~Qu, H.~Qi, and C.~Kwan, ``Unsupervised sparse dirichlet-net for
  hyperspectral image super-resolution,'' in \emph{Proc. IEEE/CVF Conference on
  Computer vision and Pattern Recognition}, 2018, pp. 2511--2520.

\bibitem{Zhu2021Hyperspectral}
Z.~{Zhu}, J.~{Hou}, J.~{Chen}, H.~{Zeng}, and J.~{Zhou}, ``Hyperspectral image
  super-resolution via deep progressive zero-centric residual learning,''
  \emph{IEEE Transactions on Image Processing}, vol.~30, pp. 1423--1438, 2021.

\bibitem{zhu2021deep}
Z.~Zhu, H.~Liu, J.~Hou, S.~Jia, and Q.~Zhang, ``Deep amended gradient descent
  for efficient spectral reconstruction from single rgb images,'' \emph{IEEE
  Transactions on Computational Imaging}, vol.~7, pp. 1176--1188, 2021.

\bibitem{zhu2021semantic}
Z.~Zhu, H.~Liu, J.~Hou, H.~Zeng, and Q.~Zhang, ``Semantic-embedded unsupervised
  spectral reconstruction from single rgb images in the wild,'' in
  \emph{Proceedings of the IEEE/CVF International Conference on Computer
  Vision}, 2021, pp. 2279--2288.

\bibitem{Xie2018Rethinking}
S.~Xie, C.~Sun, J.~Huang, Z.~Tu, and K.~Murphy, ``Rethinking spatiotemporal
  feature learning: Speed-accuracy trade-offs in video classification,'' in
  \emph{Proc. European Conference on Computer Vision}, 2018, pp. 305--321.

\bibitem{Wang2017Hyperspectral}
Y.~Wang, X.~Chen, Z.~Han, and S.~He, ``Hyperspectral image superresolution via
  nonlocal low-rank tensor approximation and total variation regularization,''
  \emph{Remote Sensing}, vol.~9, no.~12, 2017.

\bibitem{huang2014super}
H.~Huang, J.~Yu, and W.~Sun, ``Super-resolution mapping via multi-dictionary
  based sparse representation,'' in \emph{Proc. IEEE International Conference
  on Acoustics, Speech and Signal Processing}, 2014, pp. 3523--3527.

\bibitem{zhang2012super}
H.~Zhang, L.~Zhang, and H.~Shen, ``A super-resolution reconstruction algorithm
  for hyperspectral images,'' \emph{Signal Processing}, vol.~92, no.~9, pp.
  2082--2096, 2012.

\bibitem{Yuan2017Hyperspectral}
Y.~{Yuan}, X.~{Zheng}, and X.~{Lu}, ``Hyperspectral image superresolution by
  transfer learning,'' \emph{IEEE Journal of Selected Topics in Applied Earth
  Observations and Remote Sensing}, vol.~10, no.~5, pp. 1963--1974, 2017.

\bibitem{Li2018Single}
Y.~{Li}, L.~{Zhang}, C.~{Dingl}, W.~{Wei}, and Y.~{Zhang}, ``Single
  hyperspectral image super-resolution with grouped deep recursive residual
  network,'' in \emph{Proc. IEEE Fourth International Conference on Multimedia
  Big Data}, 2018, pp. 1--4.

\bibitem{Hu2020Hyperspectral}
J.~Hu, X.~Jia, Y.~Li, G.~He, and M.~Zhao, ``Hyperspectral image
  super-resolution via intrafusion network,'' \emph{IEEE Transactions on
  Geoscience and Remote Sensing}, vol.~58, no.~10, pp. 7459--7471, 2020.

\bibitem{Akhtar2016Hierarchical}
N.~Akhtar, F.~Shafait, and A.~Mian, ``Hierarchical beta process with gaussian
  process prior for hyperspectral image super resolution,'' in \emph{Proc.
  European Conference on Computer Vision}, 2016, pp. 103--120.

\bibitem{Lanaras2015Hyperspectral}
C.~Lanaras, E.~Baltsavias, and K.~Schindler, ``Hyperspectral super-resolution
  by coupled spectral unmixing,'' in \emph{Proc. IEEE/CVF International
  Conference on Computer Vision}, 2015, pp. 3586--3594.

\bibitem{Akhtar2014Sparse}
N.~Akhtar, F.~Shafait, and A.~Mian, ``Sparse spatio-spectral representation for
  hyperspectral image super-resolution,'' in \emph{Proc. European Conference on
  Computer Vision}, 2014, pp. 63--78.

\bibitem{Wang2019Deep}
W.~Wang, W.~Zeng, Y.~Huang, X.~Ding, and J.~Paisley, ``Deep blind hyperspectral
  image fusion,'' in \emph{Proc. IEEE/CVF International Conference on Computer
  Vision}, 2019, pp. 4149--4158.

\bibitem{Zhang2020Unsupervised}
L.~Zhang, J.~Nie, W.~Wei, Y.~Zhang, S.~Liao, and L.~Shao, ``Unsupervised
  adaptation learning for hyperspectral imagery super-resolution,'' in
  \emph{Proc. IEEE/CVF Conference on Computer Vision and Pattern Recognition},
  2020, pp. 3070--3079.

\bibitem{Qu2022Unsupervised}
Y.~Qu, H.~Qi, C.~Kwan, N.~Yokoya, and J.~Chanussot, ``Unsupervised and
  unregistered hyperspectral image super-resolution with mutual
  dirichlet-net,'' \emph{IEEE Transactions on Geoscience and Remote Sensing},
  vol.~60, pp. 1--18, 2022.

\bibitem{romano2015boosting}
Y.~Romano and M.~Elad, ``Boosting of image denoising algorithms,'' \emph{SIAM
  Journal on Imaging Sciences}, vol.~8, no.~2, pp. 1187--1219, 2015.

\bibitem{Tao2017Zero}
X.~Tao, C.~Zhou, X.~Shen, J.~Wang, and J.~Jia, ``Zero-order reverse
  filtering,'' in \emph{Proc. IEEE/CVF International Conference on Computer
  Vision}, 2017, pp. 222--230.

\bibitem{dong2019deep}
W.~Dong, H.~Wang, F.~Wu, G.~Shi, and X.~Li, ``Deep spatial--spectral
  representation learning for hyperspectral image denoising,'' \emph{IEEE
  Transactions on Computational Imaging}, vol.~5, no.~4, pp. 635--648, 2019.

\bibitem{wang2020spatial}
Q.~Wang, Q.~Li, and X.~Li, ``Spatial-spectral residual network for
  hyperspectral image super-resolution,'' \emph{arXiv preprint
  arXiv:2001.04609}, 2020.

\bibitem{gal2016dropout}
Y.~Gal and Z.~Ghahramani, ``Dropout as a bayesian approximation: Representing
  model uncertainty in deep learning,'' in \emph{Proc. International Conference
  on Machine Learning}, 2016, pp. 1050--1059.

\bibitem{jang2016categorical}
E.~Jang, S.~Gu, and B.~Poole, ``Categorical reparameterization with
  gumbel-softmax,'' in \emph{Proc. International Conference on Learning
  Representations (ICLR)}, 2016, pp. 1--12.

\bibitem{Kim2016Accurate}
J.~{Kim}, J.~K. {Lee}, and K.~M. {Lee}, ``Accurate image super-resolution using
  very deep convolutional networks,'' in \emph{Proc. IEEE/CVF Conference on
  Computer Vision and Pattern Recognition}, 2016, pp. 1646--1654.

\bibitem{Lim2017Enhanced}
B.~{Lim}, S.~{Son}, H.~{Kim}, S.~{Nah}, and K.~M. {Lee}, ``Enhanced deep
  residual networks for single image super-resolution,'' in \emph{Proc.
  IEEE/CVF Conference on Computer Vision and Pattern Recognition Workshops},
  2017, pp. 1132--1140.

\bibitem{Zhang2018Image}
Y.~Zhang, K.~Li, K.~Li, L.~Wang, B.~Zhong, and Y.~Fu, ``Image super-resolution
  using very deep residual channel attention networks,'' in \emph{Proc.
  European Conference on Computer Vision}, 2018, pp. 286--301.

\bibitem{Dai2019Second}
T.~{Dai}, J.~{Cai}, Y.~{Zhang}, S.~{Xia}, and L.~{Zhang}, ``Second-order
  attention network for single image super-resolution,'' in \emph{Proc.
  IEEE/CVF Conference on Computer Vision and Pattern Recognition}, 2019, pp.
  11\,057--11\,066.

\bibitem{Yasuma2010CAVE}
F.~{Yasuma}, T.~{Mitsunaga}, D.~{Iso}, and S.~K. {Nayar}, ``Generalized
  assorted pixel camera: Postcapture control of resolution, dynamic range, and
  spectrum,'' \emph{IEEE Transactions on Image Processing}, vol.~19, no.~9, pp.
  2241--2253, 2010.

\bibitem{Chakrabarti2011Harvard}
A.~{Chakrabarti} and T.~{Zickler}, ``Statistics of real-world hyperspectral
  images,'' in \emph{Proc. IEEE/CVF Conference on Computer Vision and Pattern
  Recognition}, 2011, pp. 193--200.

\bibitem{xu2019advanced}
Y.~Xu, B.~Du, L.~Zhang, D.~Cerra, M.~Pato, E.~Carmona, S.~Prasad, N.~Yokoya,
  R.~H{\"a}nsch, and B.~Le~Saux, ``Advanced multi-sensor optical remote sensing
  for urban land use and land cover classification: Outcome of the 2018 ieee
  grss data fusion contest,'' \emph{IEEE Journal of Selected Topics in Applied
  Earth Observations and Remote Sensing}, vol.~12, no.~6, pp. 1709--1724, 2019.

\bibitem{kingma2014adam}
D.~P. Kingma and J.~Ba, ``Adam: A method for stochastic optimization,''
  \emph{arXiv preprint arXiv:1412.6980}, 2014.

\bibitem{Zhou2002SSIM}
Z.~Wang and A.~Bovik, ``A universal image quality index,'' \emph{IEEE Signal
  Processing Letters}, vol.~9, no.~3, pp. 81--84, 2002.

\bibitem{Yuhas1992Discrimination}
R.~H. Yuhas, A.~F. Goetz, and J.~W. Boardman, ``Discrimination among semi-arid
  landscape endmembers using the spectral angle mapper (sam) algorithm,'' in
  \emph{Proc. Summaries 3rd Annu. JPL Airborne Geosci. Workshop}, vol.~1, 1992,
  pp. 147--149.

\bibitem{jiang2013space}
J.~Jiang, D.~Liu, J.~Gu, and S.~S{\"u}sstrunk, ``What is the space of spectral
  sensitivity functions for digital color cameras?'' in \emph{Proc. IEEE
  Workshop on Applications of Computer Vision}.\hskip 1em plus 0.5em minus
  0.4em\relax IEEE, 2013, pp. 168--179.

\bibitem{mou2017deep}
L.~Mou, P.~Ghamisi, and X.~X. Zhu, ``Deep recurrent neural networks for
  hyperspectral image classification,'' \emph{IEEE Transactions on Geoscience
  and Remote Sensing}, vol.~55, no.~7, pp. 3639--3655, 2017.

\bibitem{hong2020graph}
D.~Hong, L.~Gao, J.~Yao, B.~Zhang, A.~Plaza, and J.~Chanussot, ``Graph
  convolutional networks for hyperspectral image classification,'' \emph{IEEE
  Transactions on Geoscience and Remote Sensing}, 2020.

\bibitem{liang2018material}
J.~Liang, J.~Zhou, L.~Tong, X.~Bai, and B.~Wang, ``Material based salient
  object detection from hyperspectral images,'' \emph{Pattern Recognition},
  vol.~76, pp. 476--490, 2018.

\bibitem{zhang2018salient}
L.~Zhang, Y.~Zhang, H.~Yan, Y.~Gao, and W.~Wei, ``Salient object detection in
  hyperspectral imagery using multi-scale spectral-spatial gradient,''
  \emph{Neurocomputing}, vol. 291, pp. 215--225, 2018.

\bibitem{uzkent2017aerial}
B.~Uzkent, A.~Rangnekar, and M.~Hoffman, ``Aerial vehicle tracking by adaptive
  fusion of hyperspectral likelihood maps,'' in \emph{Proc. IEEE/CVF Conference
  on Computer Vision and Pattern Recognition Workshops}, 2017, pp. 39--48.

\bibitem{tochon2017object}
G.~Tochon, J.~Chanussot, M.~Dalla~Mura, and A.~L. Bertozzi, ``Object tracking
  by hierarchical decomposition of hyperspectral video sequences: Application
  to chemical gas plume tracking,'' \emph{IEEE Transactions on Geoscience and
  Remote Sensing}, vol.~55, no.~8, pp. 4567--4585, 2017.

\bibitem{Liu2019DARTS}
H.~Liu, K.~Simonyan, and Y.~Yang, ``Darts: Differentiable architecture
  search,'' in \emph{Proc. International Conference on Learning Representations
  (ICLR)}, 2019, pp. 1--13.

\end{thebibliography}

\begin{IEEEbiography}[{\includegraphics[width=1in,height=1.25in,clip,keepaspectratio]{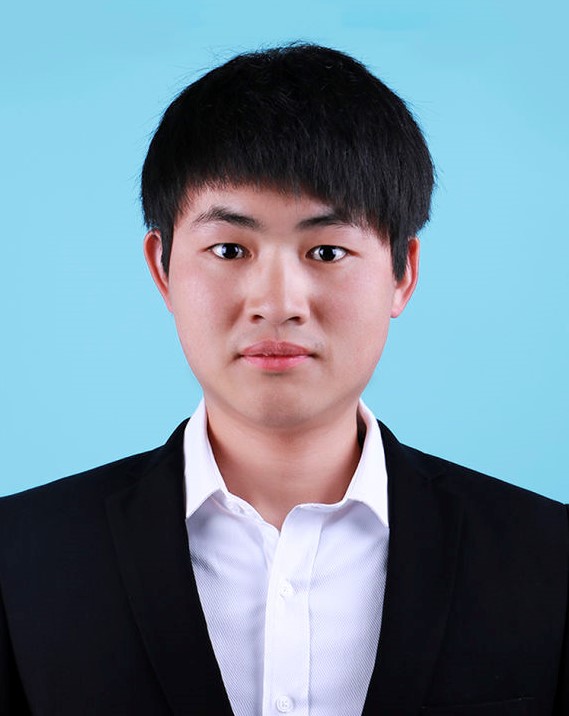}}]{Jinhui Hou} received the B.E. and M.E. degrees in communication engineering from Huaqiao University, Xiamen, China, in 2017 and 2020, respectively. He is currently pursuing the Ph.D. degree in computer science with the City University of Hong Kong. His research interests include hyperspectral image processing and deep learning.
\end{IEEEbiography}

\begin{IEEEbiography}[{\includegraphics[width=1in,height=1.25in,clip,keepaspectratio]{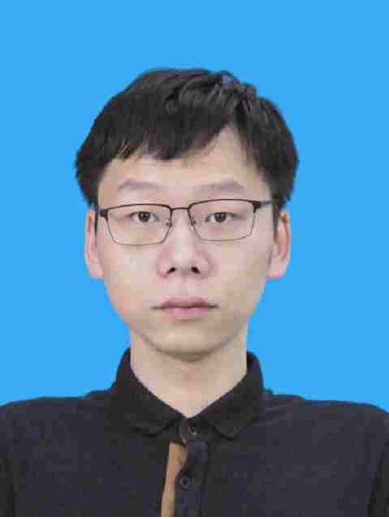}}]{Zhiyu Zhu} received the B.E. and M.E. degrees in Mechatronic Engineering, both from Harbin Institute of Technology, in 2017 and 2019, respectively. He is currently pursuing the Ph.D. degree in computer science with the City University of Hong Kong. His research interests include hyperspectral image processing and deep learning.
\end{IEEEbiography}
\vspace{-14mm}

\begin{IEEEbiography}[{\includegraphics[width=1in,height=1.25in,clip,keepaspectratio]{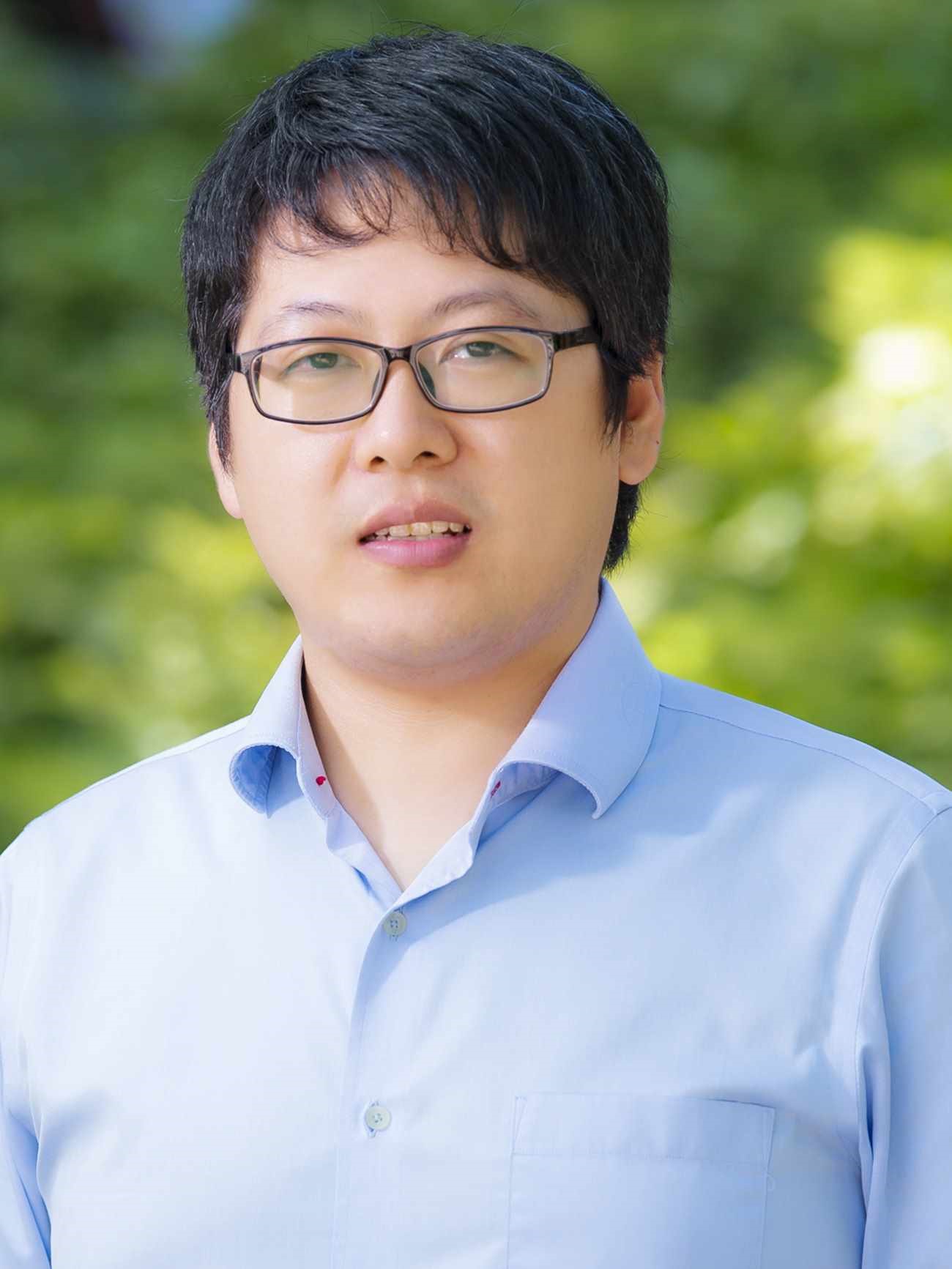}}]{Junhui Hou} (Senior Member, IEEE) is an Assistant Professor with the Department of Computer Science, City University of Hong Kong. He received the B.Eng. degree in information engineering (Talented Students Program) from the South China University of Technology, Guangzhou, China, in 2009, the M.Eng. degree in signal and information processing from Northwestern Polytechnical University, Xian, China, in 2012, and the Ph.D. degree in electrical and electronic engineering from the School of Electrical and Electronic Engineering, Nanyang Technological University, Singapore, in 2016. His research interests fall into the general areas of multimedia signal processing, such as image/video/3D geometry data representation, processing and analysis, graph-based clustering/classification, and data compression.

He received the Chinese Government Award for Outstanding Students Study Abroad from China Scholarship Council in 2015 and the Early Career Award (3/381) from the Hong Kong Research Grants Council in 2018. He is an elected member of MSA-TC, VSPC-TC, and MMSP-TC. He is currently an Associate Editor for IEEE Transactions on Image Processing, IEEE Transactions on Circuits and Systems for Video Technology, Signal Processing: Image Communication, and The Visual Computer. He also served as the Guest Editor for the IEEE Journal of Selected Topics in Applied Earth Observations and Remote Sensing and as an Area Chair of ACM MM'19/20/21/22, IEEE ICME'20, VCIP'20/21/22, ICIP'22, and WACV'21.
\end{IEEEbiography}
\vspace{-1.3cm}

\begin{IEEEbiography}[{\includegraphics[width=1in,height=1.25in,clip,keepaspectratio]{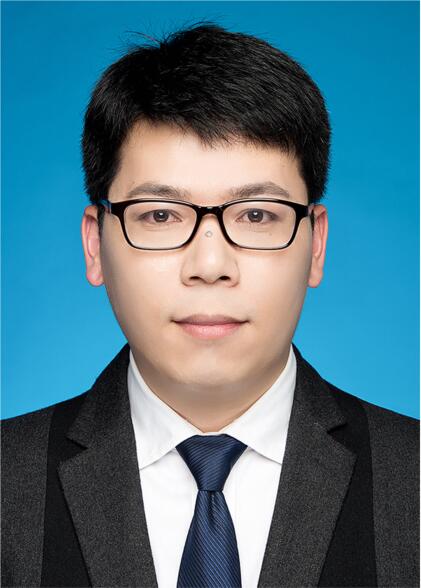}}]{Huanqiang Zeng} (Senior Member, IEEE)
received the B.S. and M.S. degrees in electrical engineering from Huaqiao University, China, and the Ph.D. degree in electrical engineering from Nanyang Technological University, Singapore.

He is currently a Full Professor at the School of Engineering and the School of Information Science and Engineering, Huaqiao University. Before that, he was a Postdoctoral Fellow at The Chinese University of Hong Kong, Hong Kong. He has published more than 100 papers in well-known journals and conferences, including three best poster/paper awards (in the International Forum of Digital TV and Multimedia Communication 2018 and the Chinese Conference on Signal Processing 2017/2019). His research interests include image processing, video coding, machine learning, and computer vision. He has also been actively serving as the General Co-Chair for IEEE International Symposium on Intelligent Signal Processing and Communication Systems 2017 (ISPACS2017), the Co-Organizer for ICME2020 Workshop on 3D Point Cloud Processing, Analysis, Compression, and Communication, the Technical Program Co-Chair for Asia–Pacific Signal and Information Processing Association Annual Summit and Conference 2017 (APSIPA-ASC2017), the Area Chair for IEEE International Conference on Visual Communications and Image Processing (VCIP2015 and VCIP2020), and a technical program committee member for multiple flagship international conferences. He has been actively serving as an Associate Editor for IEEE Transactions on Image Processing, IEEE Transactions on Circuits and Systems for Video Technology, and Electronics Letters (IET). He has been actively serving as a Guest Editor for Journal of Visual Communication and Image Representation, Multimedia Tools and Applications, and Journal of Ambient Intelligence and Humanized Computing.
\end{IEEEbiography}
\vspace{-10mm}

\begin{IEEEbiography}[{\includegraphics[width=1in,height=1.25in,clip,keepaspectratio]{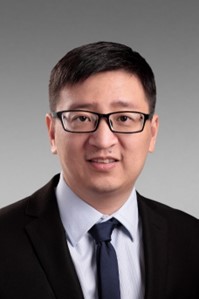}}]{Jinjian Wu} (Member, IEEE) received the B.Sc. and Ph.D. degrees from Xidian University, Xi'an, China, in 2008 and 2013, respectively. From 2011 to 2013, he was a Research Assistant with Nanyang Technological University, Singapore, where he was a Postdoctoral Research Fellow from 2013 to 2014. From 2015 to 2019, he was an Associate Professor with Xidian University, where he has been a Professor since 2019. His research interests include visual perceptual modeling, biomimetic imaging, quality evaluation, and object detection. He received the Best Student Paper Award/candidate at the ISCAS 2013/CICAI 2021. He has served as an Associate Editor for the journal of Circuits, Systems and Signal Processing (CSSP), the Special Section Chair for the IEEE Visual Communications and Image Processing (VCIP) 2017, and the Section Chair/Organizer/TPC Member for the ICME 2014-2015, PCM 2015-2016, VCIP 2018, AAAI 2019-2021, and ACM MM 2021-2022.
\end{IEEEbiography}
\vspace{-25mm}

\begin{IEEEbiography}[{\includegraphics[width=1in,height=1.25in,clip,keepaspectratio]{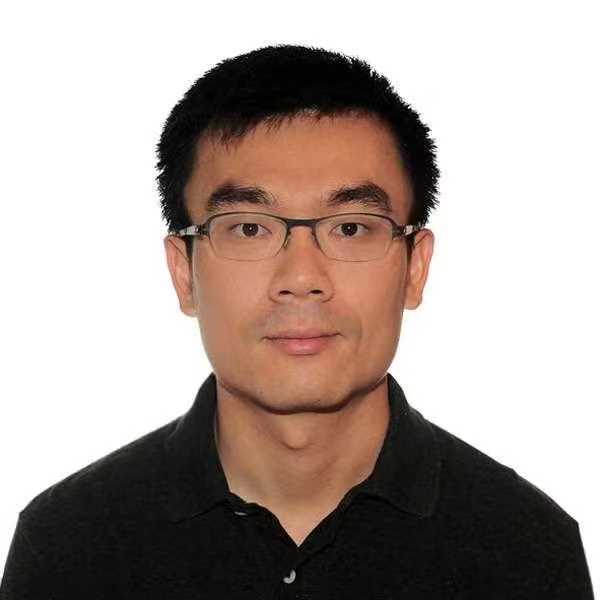}}]{Jiantao Zhou} (Senior Member, IEEE) received the B.Eng. degree from the Department of Electronic Engineering, Dalian University of Technology, in 2002, the M.Phil. degree from the Department of Radio Engineering, Southeast University, in 2005, and the Ph.D. degree from the Department of Electronic and Computer Engineering, Hong Kong University of Science and Technology, in 2009. He held various research positions with the University of Illinois at Urbana-Champaign, Hong Kong University of Science and Technology, and McMaster University. He is an Associate Professor with the Department of Computer and Information Science, Faculty of Science and Technology, University of Macau, and also the Interim Head of the newly established Centre for Artificial Intelligence and Robotics. His research interests include multimedia security and forensics, multimedia signal processing, artificial intelligence, and big data. He holds four granted U.S. patents and two granted Chinese patents. He has coauthored two papers that received the Best Paper Award at the IEEE Pacific-Rim Conference on Multimedia in 2007 and the Best Student Paper Award at the IEEE International Conference on Multimedia and Expo in 2016. He is serving as an Associate Editor for the IEEE TRANSACTIONS ON IMAGE PROCESSING and the IEEE TRANSACTIONS ON MULTIMEDIA.
\end{IEEEbiography}

\end{document}